\documentclass[11pt,a4paper]{article}
\usepackage{times,latexsym}
\usepackage{url}
\usepackage[T1]{fontenc}

\usepackage{graphicx}
\graphicspath{{images/}}

\usepackage[acceptedWithA]{tacl2018v2}
\usepackage{xcolor}
\usepackage{soul}
\definecolor{mycol}{rgb}{0.549, 0.794, 1.0}
\DeclareRobustCommand{\hlcustom}[1]{{\sethlcolor{mycol}\hl{#1}}}  % change highlight color

\usepackage{xspace,mfirstuc,tabulary}

\newif\iftaclinstructions
\taclinstructionsfalse % AUTHORS: do NOT set this to true
\iftaclinstructions
\renewcommand{\confidential}{}
\renewcommand{\anonsubtext}{(No author info supplied here, for consistency with
TACL-submission anonymization requirements)}
\newcommand{\instr}
\fi

\iftaclpubformat % this "if" is set by the choice of options

\else

\fi

\newcommand{\dataset}[1]{\ensuremath{\mathcal{D_{\mathrm{#1}}}}}
\newcommand{\datasetsplit}[2]{\dataset{#1}$^{#2}$}
\newcommand{\squad}{SQuAD}
\newcommand{\squadone}{SQuAD1.1}

\newcommand{\swag}{SWAG}
\newcommand{\hotpot}{HotpotQA}
\newcommand{\drop}{DROP}
\newcommand{\naturalquestions}{NaturalQuestions}

\newcommand{\std}[1]{$_{\text{\thinspace{#1}}}$}
\newcommand{\resultsemph}[1]{\underline{#1}}

\usepackage{booktabs}
\usepackage{amsmath}
\usepackage{tabularx}
\usepackage{multirow}
\usepackage{colortbl}
\usepackage{hhline}

\definecolor{CustomGray}{gray}{0.95}
\newcolumntype{a}{>{\columncolor{CustomGray}}c}

\usepackage{adjustbox}
\newcolumntype{R}[2]{%
    >{\adjustbox{angle=#1,lap=\width-(#2)}\bgroup}%
    l%
    <{\egroup}%
}
\newif\ifpagebreaks
\pagebreakstrue
\newif\ifarxiv
\arxivfalse

\newif\ifappendix
\appendixtrue

\ifarxiv
    \appendixtrue
\fi

\title{Beat the AI: Investigating Adversarial Human Annotation\\for Reading Comprehension}

\author{
  Max Bartolo \quad Alastair Roberts \quad Johannes Welbl \quad Sebastian Riedel \quad Pontus Stenetorp \\
  Department of Computer Science \\
  University College London \\
  {\tt \{m.bartolo,a.roberts,j.welbl,s.riedel,p.stenetorp\}@cs.ucl.ac.uk}
}

\date{}

\begin{document}
\maketitle

%%%%%%%%%%%%%%%%%%%%%%%%%%%%%%%%%%%%%%%%%%%%%%%%%%
\begin{abstract}
Innovations in annotation methodology have been a catalyst for Reading Comprehension (RC) datasets and models.
One recent trend to challenge current RC models is to involve a model in the annotation process: humans create questions adversarially, such that the model fails to answer them correctly.
In this work we investigate this annotation methodology and apply it in three different settings, collecting a total of 36,000 samples with progressively stronger models in the annotation loop.
This allows us to explore questions such as the reproducibility of the adversarial effect, transfer from data collected with varying model-in-the-loop strengths, and generalisation to data collected without a model.
We find that training on adversarially collected samples leads to strong generalisation to non-adversarially collected datasets, yet with progressive performance deterioration with increasingly stronger models-in-the-loop.
Furthermore, we find that stronger models can still learn from datasets collected with substantially weaker models-in-the-loop.
When trained on data collected with a BiDAF model in the loop, RoBERTa achieves 39.9F$_\text{1}$ on questions that it cannot answer when trained on \squad{} -- only marginally lower than when trained on data collected using RoBERTa itself~(41.0F$_\text{1}$).
\end{abstract}
%%%%%%%%%%%%%%%%%%%%%%%%%%%%%%%%%%%%%%%%%%%%%%%%%%

%%%%%%%%%%%%%%%%%%%%%%%%%%%%%%
%%%   PROGRESSION FIGURE  %%%%
%%%%%%%%%%%%%%%%%%%%%%%%%%%%%%
\begin{figure}[t]
    \centering
    \includegraphics[width=\columnwidth]{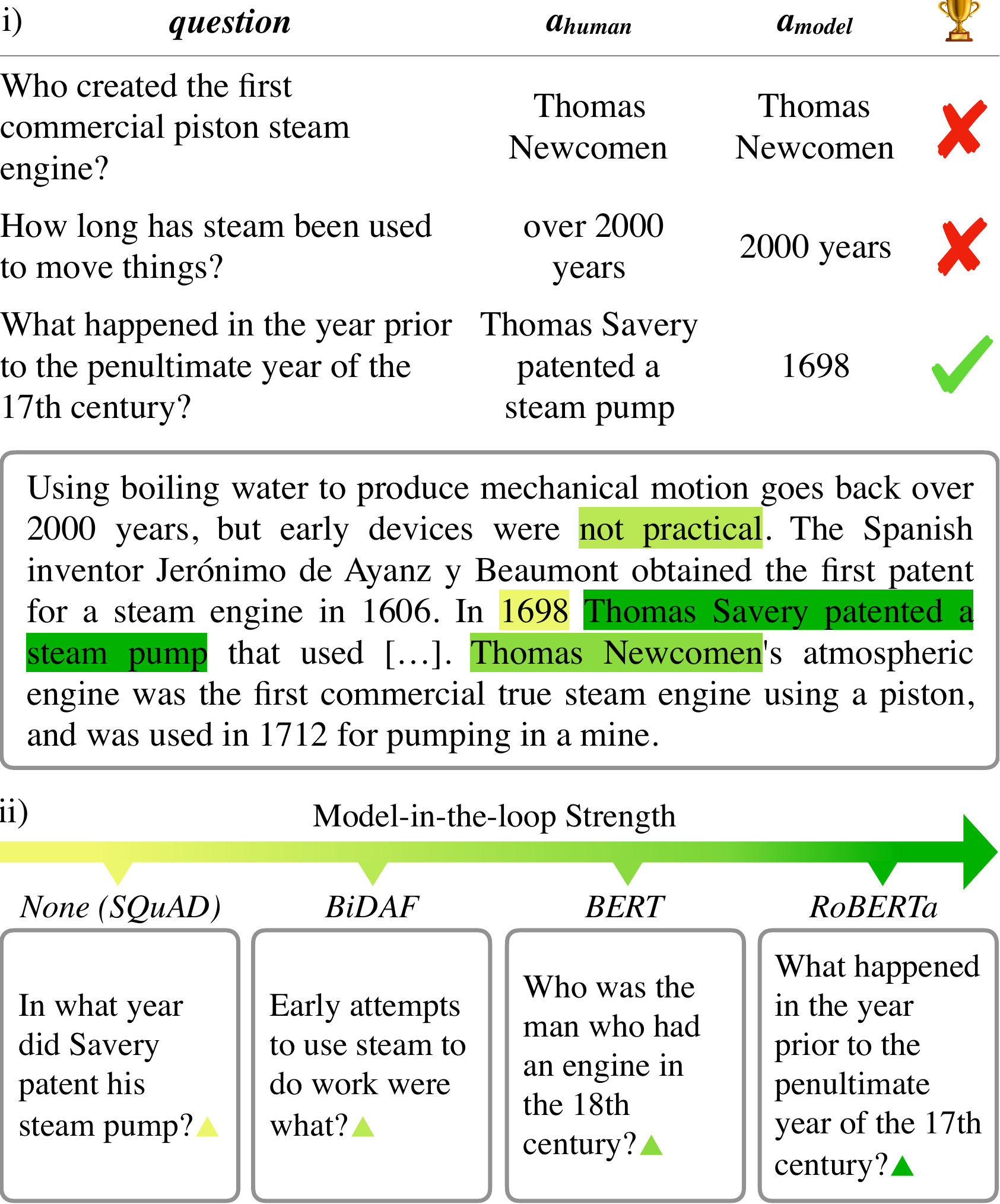}
    \caption{Human annotation with a model in the loop, showing: i) the ``Beat the AI'' annotation setting where only questions that the model does not answer correctly are accepted, and ii) questions generated this way, with a progressively stronger model in the annotation loop.
    } 
    \label{fig:progression}
\end{figure}
%%%%%%%%%%%%%%%%%%%%%%%%%%%%%%

\section{Introduction}

Data collection is a fundamental prerequisite for Machine Learning-based approaches to Natural Language Processing (NLP).
Innovations in data acquisition methodology, such as crowdsourcing, have led to major breakthroughs in scalability and preceded the ``deep learning revolution'', for which they can arguably be seen as co-responsible~\cite{deng2009imagenet,bowman-etal-2015-large,rajpurkar2016squad}.
Annotation approaches include expert annotation, for example,~relying on trained linguists~\cite{marcus1993penntreebank}, crowd-sourcing by non-experts~\cite{snow2008cheap}, distant supervision~\cite{mintz2009distant, joshi-etal-2017-triviaqa},
and leveraging document structure~\cite{herman2015teaching}.
The concrete data collection paradigm chosen dictates the degree of scalability, annotation cost, precise task structure (often arising as a compromise of the above) and difficulty, domain coverage, as well as resulting dataset biases and model blind spots~\cite{jia2017adversarial,schwartz2017effect,gururangan2018annotation}.

A recently emerging trend in NLP dataset creation is the use of a \emph{model-in-the-loop} when composing samples:
A contemporary model is used either as a filter or directly during annotation, to identify samples wrongly predicted by the model.
Examples of this method are realised in \emph{Build It Break It, The Language Edition}~\cite{ettinger2017buildit}, \hotpot{}~\cite{yang2018hotpotqa}, \swag{}~\cite{zellers2018swag}, Mechanical Turker Descent~\cite{yang2018mastering}, \drop{}~\cite{dua2019drop}, CODAH~\cite{chen-etal-2019-codah}, Quoref~\cite{dasigi-etal-2019-quoref}, and AdversarialNLI~\cite{nie2019adversarial}.\footnote{
    The idea was alluded to at least as early as~\citet{richardson2013mctest}, but it has only recently seen wider adoption.
}
This approach probes model robustness and ensures that the resulting datasets pose a challenge to current models, which drives research to tackle new sets of problems.
We study this approach in the context of RC, and investigate its robustness in the face of continuously progressing models -- do adversarially constructed datasets quickly become outdated in their usefulness as models grow stronger? 
Based on models trained on the widely used \squad{} dataset, and following the same annotation protocol, we investigate the annotation setup where an annotator has to compose questions for which the model predicts the wrong answer.
As a result, only samples that the model fails to predict correctly are retained in the dataset -- see Figure~\ref{fig:progression} for an example.

We apply this annotation strategy with three distinct models in the loop, resulting in datasets with 12,000 samples each.
We then study the reproducibility of the adversarial effect when retraining the models with the same data, as well as the generalisation ability of models trained using datasets produced with and without a model adversary.
Models can, to a considerable degree, learn to generalise to more challenging questions, based on training sets collected with both stronger and also weaker models in the loop.
Compared to training on \squad{}, training on adversarially composed questions leads to a similar degree of generalisation to non-adversarially written questions, both for \squad{} and \naturalquestions{}~\cite{kwiatkowski-etal-2019-natural}.
It furthermore leads to general improvements across the model-in-the-loop datasets we collect, as well as improvements of more than 20.0F$_\text{1}$ for both BERT and RoBERTa on an extractive subset of \drop{}~\cite{dua2019drop}, another adversarially composed dataset.
When conducting a systematic analysis of the concrete questions different models fail to answer correctly, as well as non-adversarially composed questions, we see that the nature of the resulting questions changes:
Questions composed with a model in the loop are overall more diverse, use more paraphrasing, multi-hop inference, comparisons, and background knowledge, and are generally less easily answered by matching an explicit statement that states the required information literally.
Given our observations, we believe a model-in-the-loop approach to annotation shows promise and should be considered when creating future RC datasets.

To summarise, our contributions are as follows:
First, an investigation into the model-in-the-loop approach to RC data collection based on three progressively stronger models, together with an empirical performance comparison when trained on datasets constructed with adversaries of different strength.
Second, a comparative investigation into the nature of questions composed to be unsolvable by a sequence of progressively stronger models.
Third, a study of the reproducibility of the adversarial effect and the generalisation ability of models trained in various settings.

\section{Related Work}

\paragraph{Constructing Challenging Datasets}{
Recent efforts in dataset construction have driven considerable progress in RC, yet datasets are structurally diverse and annotation methodologies vary.
With its large size and combination of free-form questions with answers as extracted spans, \squadone{}~\cite{rajpurkar2016squad} has become an established benchmark that has inspired the construction of a series of similarly structured datasets. 
However, mounting evidence suggests that models can achieve strong generalisation performance merely by relying on superficial cues -- such as lexical overlap, term frequencies, or entity type matching~\cite{chen-etal-2016-thorough, weissenborn-etal-2017-making, sugawara-etal-2018-makes}.
It has thus become an increasingly important consideration to construct datasets that RC models find challenging, and for which natural language understanding is a requisite for generalisation.
Attempts to achieve this non-trivial aim have typically revolved around extensions to the \squad{} dataset annotation methodology. 
They include unanswerable questions~\cite{trischler-etal-2017-newsqa, rajpurkar-etal-2018-know, reddy-etal-2019-coqa, choi-etal-2018-quac}, adding the option of ``Yes'' or ``No'' answers~\cite{dua2019drop, kwiatkowski-etal-2019-natural}, questions requiring reasoning over multiple sentences or documents~\cite{welbl2018constructing, yang2018hotpotqa}, questions requiring rule interpretation or context awareness~\cite{saeidi-etal-2018-interpretation, choi-etal-2018-quac, reddy-etal-2019-coqa}, limiting annotator passage exposure by sourcing questions first~\cite{kwiatkowski-etal-2019-natural}, controlling answer types by including options for dates, numbers, or spans from the question~\cite{dua2019drop}, as well as questions with free form answers~\cite{nguyen2016ms, kocisky2018narrativeqa, reddy-etal-2019-coqa}.
}

\paragraph{Adversarial Annotation}{
One recently adopted approach to constructing challenging datasets involves the use of an adversarial model to select examples that it does not perform well on, an approach which superficially is akin to active learning~\cite{lewis1994sequential}.
Here, we make a distinction between two sub-categories of adversarial annotation:
i) \emph{adversarial filtering}, where the adversarial model is applied offline in a separate stage of the process, usually after data generation;
examples include SWAG~\cite{zellers2018swag}, ReCoRD~\cite{zhang2018record}, \hotpot{}~\cite{yang2018hotpotqa}, and HellaSWAG~\cite{zellers-etal-2019-hellaswag};
ii)~\emph{model-in-the-loop adversarial annotation}, where the annotator can directly interact with the adversary during the annotation process and uses the feedback to further inform the generation process;
examples include CODAH~\cite{chen-etal-2019-codah}, Quoref~\cite{dasigi-etal-2019-quoref}, DROP~\cite{dua2019drop}, FEVER2.0~\cite{thorne-etal-2019-fever2}, AdversarialNLI~\cite{nie2019adversarial}, as well as work by~\citet{dinan-etal-2019-build}, ~\citet{Kaushik2020Learning}, and~\citet{wallace-etal-2019-trick} for the Quizbowl task. 

We are primarily interested in the latter category, as this feedback loop creates an environment where the annotator can probe the model directly to explore its weaknesses and formulate targeted adversarial attacks.
Although~\citet{dua2019drop} and~\citet{dasigi-etal-2019-quoref} make use of adversarial annotations for RC, both annotation setups limit the reach of the model-in-the-loop: 
In DROP, primarily due to the imposition of specific answer types, and in Quoref by focusing on co-reference, which is already a known RC model weakness.

In contrast, we investigate a scenario where annotators interact with a model in its original task setting -- annotators must thus explore a range of natural adversarial attacks, as opposed to filtering out ``easy'' samples during the annotation process.
}

\section{Annotation Methodology}

\subsection{Annotation Protocol}
The data annotation protocol is based on \squadone{}, with a model in the loop, and the additional instruction that questions should only have one answer in the passage, which directly mirrors the setting in which these models were trained.

\begin{figure}[t]
    \centering
    \includegraphics[width=\columnwidth]{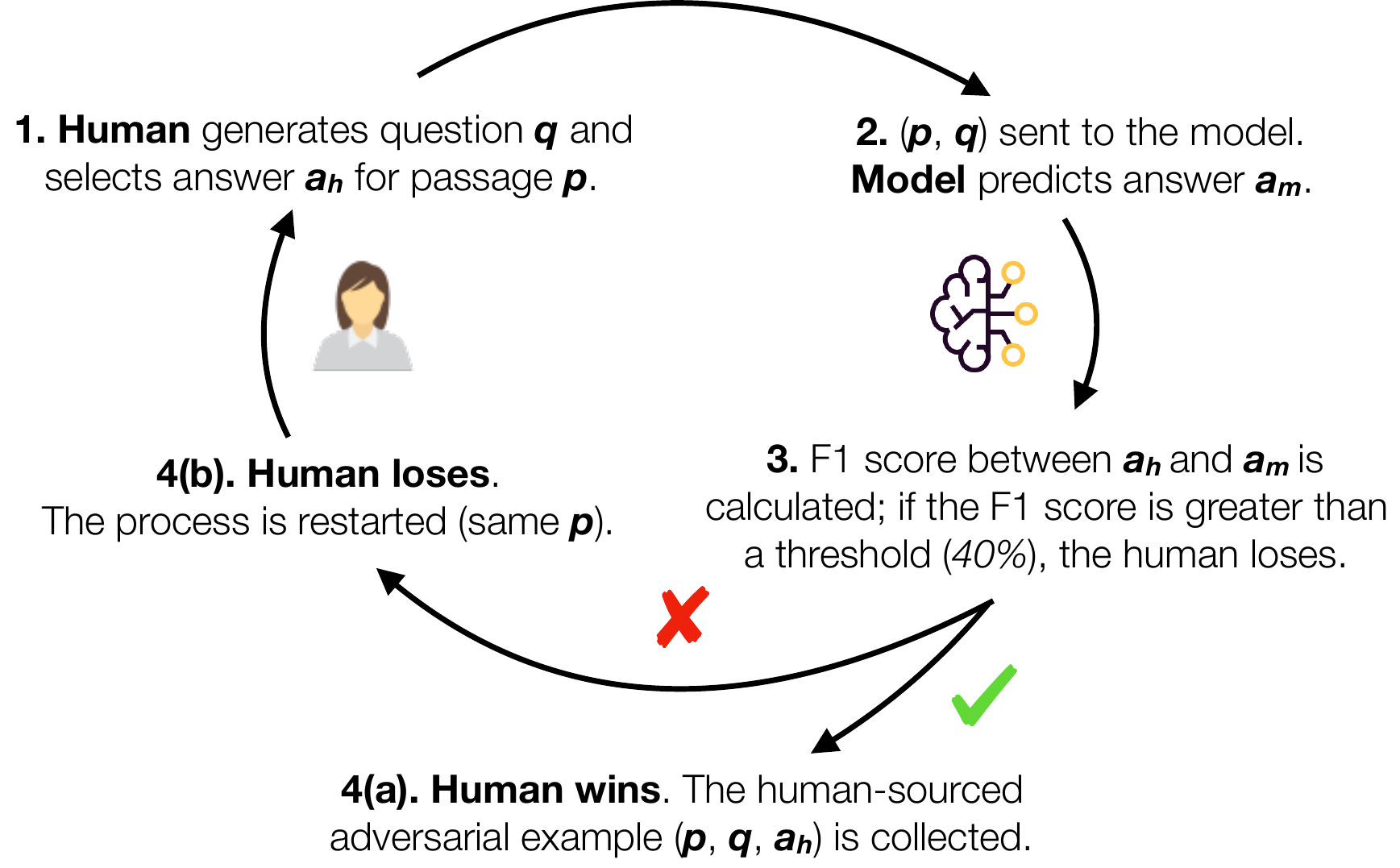}
    \caption{Overview of the annotation process to collect adversarially written questions from humans using a model in the loop.
    } 
    \label{fig:data_collection}
\end{figure}

Formally, provided with a passage $p$, a human annotator generates a question $q$ and selects a (human) answer $a_h$ by highlighting the corresponding span in the passage. 
The input $(p, q)$ is then given to the model, which returns a predicted (model) answer $a_m$. 
To compare the two, a word-overlap F$_\text{1}$ score between $a_h$ and $a_m$ is computed; 
a score above a threshold of $40\%$ is considered a ``win'' for the model.\footnote{
This threshold is set after initial experiments to not be overly restrictive given acceptable answer spans, e.g.,~a human answer of ``New York'' vs.~model answer ``New York City'' would still lead to a model ``win''.
}
This process is repeated until the human ``wins'';
Figure~\ref{fig:data_collection} gives a schematic overview of the process. 
All successful $(p, q, a_h)$ triples, that is,~those which the model is unable to answer correctly, are then retained for further validation.
%

%%%%%%%%%%%%%%%%%%%%%%%%%%%%%%
%%%   DATA COLLECTION FIGURE  %%%%
%%%%%%%%%%%%%%%%%%%%%%%%%%%%%%
%%%%%%%%%%%%%%%%%%%%%%%%%%%%%%

%%%%%%%%%%%%%%%%%%%%%%%%%%%%%%
%%%   Question Generation FIGURE  %%%%
%%%%%%%%%%%%%%%%%%%%%%%%%%%%%%
\begin{figure*}[!ht]
    \centering
    \includegraphics[width=0.99\linewidth]{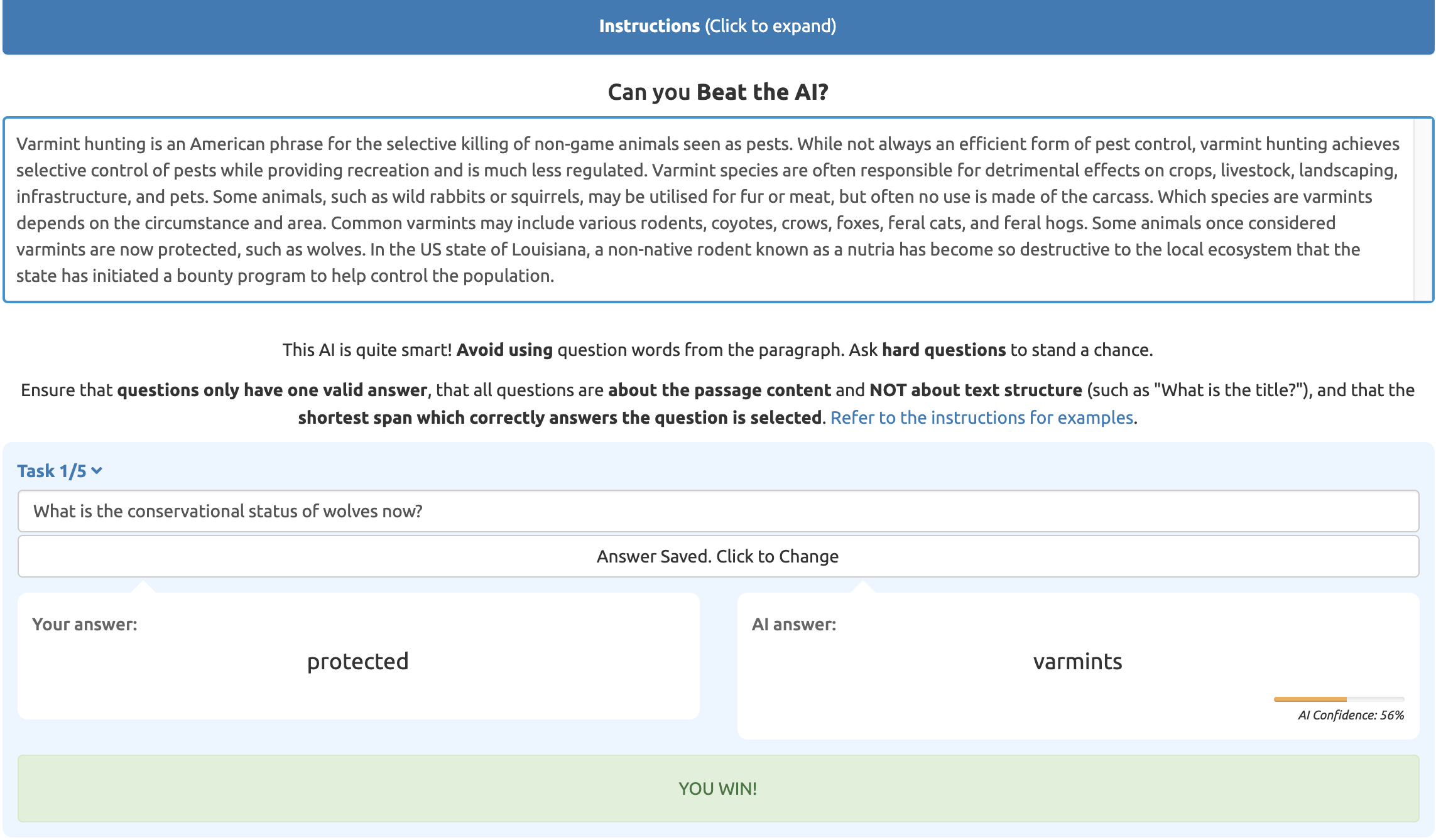}
    \caption{``Beat the AI'' question generation interface. Human annotators are tasked with asking questions about a provided passage which the model in the loop fails to answer correctly.} 
    \label{fig:interface_question_generation}
\end{figure*}
%%%%%%%%%%%%%%%%%%%%%%%%%%%%%%

\subsection{Annotation Details}
\paragraph{Models in the Annotation Loop}{
We begin by training three different models, which are used as adversaries during data annotation.
As a seed dataset for training the models we select the widely used \squadone{}~\cite{rajpurkar2016squad} dataset, a large-scale resource for which a variety of mature and well-performing models are readily available.
Furthermore, unlike cloze-based datasets, \squad{} is robust to passage/question-only adversarial attacks~\cite{kaushik-lipton-2018-much}.
We will compare dataset annotation with a series of three progressively stronger models as adversary in the loop, namely BiDAF~\cite{Seo2016BidAF}, BERT$_\text{LARGE}$~\cite{devlin2019bert}, and RoBERTa$_\text{LARGE}$~\cite{liu2019roberta}. 
Each of these will serve as a model adversary in a separate annotation experiment and result in three distinct datasets; we will refer to these as \dataset{BiDAF}, \dataset{BERT}, and \dataset{RoBERTa} respectively. Examples from the validation set of each are shown in Table~\ref{tab:dataset_examples}.
We rely on the \emph{AllenNLP}~\cite{gardner-etal-2018-allennlp} and \emph{Transformers}~\cite{Wolf2019HuggingFacesTS} model implementations, and our models achieve EM/F$_\text{1}$ scores of 65.5\%/77.5\%, 82.7\%/90.3\% and 86.9\%/93.6\% for BiDAF, BERT, and RoBERTa respectively on the \squadone{} validation set, consistent with results reported in other work.
%

%%%=================== Table of examples - START ===================%%%
\begin{table*}[!ht]
\centering \footnotesize
\begin{tabular}{p{0.005\textwidth}p{0.935\textwidth}}

 \toprule

\parbox[t]{1mm}{\multirow{2}{*}{\rotatebox[origin=c]{90}{{BiDAF}}}} &
\textbf{Passage:} \textit{[\ldots] the United Methodist Church has placed great emphasis on the importance of education. As such, the United Methodist Church established and is affiliated with around one hundred colleges [\dots] of Methodist-related Schools, Colleges, and Universities. The church operates \hlcustom{three hundred sixty schools} and institutions overseas.} \\
& \textbf{Question:} \textit{The United Methodist Church has how many schools internationally?} \\

\midrule

\parbox[t]{1mm}{\multirow{2}{*}{\rotatebox[origin=c]{90}{{BiDAF}}}} &
\textbf{Passage:} \textit{In a purely capitalist mode of production (i.e. where professional and labor organizations cannot limit the number of workers) the workers wages will not be controlled by these organizations, or by the employer, but rather by \hlcustom{the market}. Wages work in the same way as prices for any other good. Thus, wages can be considered as a [\ldots]} \\
& \textbf{Question:} \textit{What determines worker wages?} \\

\midrule

\parbox[t]{1mm}{\multirow{2}{*}{\rotatebox[origin=c]{90}{{BiDAF}}}} &
\textbf{Passage:} \textit{[\ldots] released to the atmosphere, and a separate source of water feeding the boiler is supplied. Normally \hlcustom{water} is the fluid of choice due to its favourable properties, such as non-toxic and unreactive chemistry, abundance, low cost, and its thermodynamic properties. Mercury is the working fluid in the mercury vapor turbine [\ldots]} \\
& \textbf{Question:} \textit{What is the most popular type of fluid?} \\

\midrule
\midrule

\parbox[t]{1mm}{\multirow{2}{*}{\rotatebox[origin=c]{90}{{BERT}}}} &
\textbf{Passage:} \textit{[\ldots] Jochi was secretly poisoned by an order from Genghis Khan. Rashid al-Din reports that the great Khan sent for his sons in the spring of 1223, and while \hlcustom{his brothers} heeded the order, Jochi remained in Khorasan. Juzjani suggests that the disagreement arose from a quarrel between Jochi and his brothers in the siege of Urgench [\ldots]} \\
& \textbf{Question:} \textit{Who went to Khan after his order in 1223?} \\

\midrule

\parbox[t]{1mm}{\multirow{2}{*}{\rotatebox[origin=c]{90}{{BERT}}}} &
\textbf{Passage:} \textit{In the Sandgate area, to the east of the city and beside the river, resided the close-knit community of keelmen and their families. They were so called because [\ldots] transfer coal from the river banks to the waiting colliers, for export to London and elsewhere. In the 1630s about 7,000 out of 20,000 inhabitants of \hlcustom{Newcastle} died of plague [\ldots]} \\
& \textbf{Question:} \textit{Where did almost half the people die?} \\

\midrule

\parbox[t]{1mm}{\multirow{2}{*}{\rotatebox[origin=c]{90}{{BERT}}}} &
\textbf{Passage:} \textit{[\ldots] was important to reduce the weight of coal carried. Steam engines remained the dominant source of power until the early 20th century, when \hlcustom{advances in the design of electric motors and internal combustion engines} gradually resulted in the replacement of reciprocating (piston) steam engines, with shipping in the 20th-century [\ldots]} \\
& \textbf{Question:} \textit{Why did steam engines become obsolete?} \\

\midrule
\midrule

\parbox[t]{1mm}{\multirow{2}{*}{\rotatebox[origin=c]{90}{{RoBERTa}}}} &
\textbf{Passage:} \textit{[\ldots] and seven other hymns were published in the Achtliederbuch, the first Lutheran hymnal. In 1524 Luther developed his original \hlcustom{four}-stanza psalm paraphrase into a five-stanza Reformation hymn that developed the theme of "grace alone" more fully. Because it expressed essential Reformation doctrine, this expanded version of "Aus [\ldots]} \\
& \textbf{Question:} \textit{Luther's reformed hymn did not feature stanzas of what quantity?} \\

\midrule

\parbox[t]{1mm}{\multirow{2}{*}{\rotatebox[origin=c]{90}{{RoBERTa}}}} &
\textbf{Passage:} \textit{[\ldots] tight end Greg Olsen, who caught a career-high 77 passes for 1,104 yards and seven touchdowns, and wide receiver \hlcustom{Ted Ginn, Jr.}, who caught 44 passes for 739 yards and 10 touchdowns; [\ldots] receivers included veteran Jerricho Cotchery (39 receptions for 485 yards), rookie Devin Funchess (31 receptions for 473 yards and [\ldots]} \\
& \textbf{Question:} \textit{Who caught the second most passes?} \\

\midrule

\parbox[t]{1mm}{\multirow{2}{*}{\rotatebox[origin=c]{90}{{RoBERTa}}}} &
\textbf{Passage:} \textit{Other prominent alumni include anthropologists David Graeber and Donald Johanson, who is best known for discovering the fossil of a female hominid australopithecine known as "Lucy" in the Afar Triangle region, psychologist John B. Watson, American psychologist who established the psychological school of behaviorism, communication theorist Harold Innis, chess grandmaster \hlcustom{Samuel Reshevsky}, and conservative international relations scholar and White House Coordinator of Security Planning for the National Security Council Samuel P. Huntington.} \\
& \textbf{Question:} \textit{Who thinks three moves ahead?} \\

\bottomrule
\end{tabular}

\caption{Validation set examples of questions collected using different RC models (BiDAF, BERT, and RoBERTa) in the annotation loop. The answer to the question is highlighted in the passage.}
\label{tab:dataset_examples}

\end{table*}
%%%=================== Table of examples - END ===================%%%

%
Our choice of models reflects both the transition from LSTM-based to pre-trained transformer-based models, as well as a graduation among the latter; we investigate how this is reflected in datasets collected with each of these different models in the annotation loop. 
For each of the models we collect 10,000 training, 1,000 validation, and 1,000 test examples. Dataset sizes are motivated by the data efficiency of transformer-based pretrained models~\cite{devlin2019bert,liu2019roberta}, which has improved the viability of smaller-scale data collection efforts for investigative and analysis purposes.

To ensure the experimental integrity provided by reporting all results on a held-out test set, we split the existing \squadone{} validation set in half (stratified by document title) as the official test set is not publicly available. 
We maintain passage consistency across the training, validation and test sets of all datasets to enable like-for-like comparisons.
Lastly, we use the majority vote answer as ground truth for \squadone{} to
ensure that all our datasets have one valid answer per question, enabling us to fairly draw direct comparisons.
For clarity, we will hereafter refer to this modified version of \squadone{} as \dataset{SQuAD}. 
}
\paragraph{Crowdsourcing}
We use custom-designed Human Intelligence Tasks (HITs) served through Amazon Mechanical Turk (AMT) for all annotation efforts\ifarxiv (see Appendix~\ref{sec:appendix_dataset_construction})\fi.
Workers are required to be based in Canada, the UK, or the US, have a HIT Approval Rate greater than $98\%$, and have previously completed at least 1,000 HITs successfully. 
We experiment with and without the AMT \emph{Master} requirement and find no substantial difference in quality, but observe a throughput reduction of nearly 90\%.
We pay USD 2.00 for every question generation HIT, during which workers are required to compose up to five questions that ``beat'' the model in the loop (cf. Figure~\ref{fig:interface_question_generation}). 
The mean HIT completion times for BiDAF, BERT, and RoBERTa are 551.8s, 722.4s, and 686.4s.
Furthermore we find that human workers are able to generate questions that successfully ``beat'' the model in the loop $59.4\%$ of the time for BiDAF, $47.1\%$ for BERT, and $44.0\%$ for RoBERTa.
These metrics broadly reflect the relative strength of the models.

\subsection{Quality Control}
\paragraph{Training and Qualification}{
We provide a two-part worker training interface in order to i) familiarise workers with the process, and ii) conduct a first screening based on worker outputs.
The interface familiarises workers with formulating questions, and answering them through span selection.
Workers are asked to generate questions for two given answers, to highlight answers for two given questions, to generate one full question-answer pair, and finally to complete a question generation HIT with BiDAF as the model in the loop. 
Each worker's output is then reviewed manually (by the authors); those who pass the screening are added to the pool of qualified annotators.
}

\paragraph{Manual Worker Validation}{
In the second annotation stage, qualified workers produce data for the ``Beat the AI'' question generation task.
A sample of every worker's HITs is manually reviewed based on their total number of completed tasks $n$, determined by $\lfloor 5 \cdot \log_{10}(n) + 1 \rfloor$, chosen for convenience. 
This is done after every annotation batch; if workers fall below an $80\%$ success threshold at any point, their qualification is revoked and their work is discarded in its entirety.
}

\paragraph{Question Answerability}{
As the models used in the annotation task become stronger, the resulting questions tend to become more complex.
However, this also means that it becomes more challenging to disentangle measures of dataset quality from inherent question difficulty. 
As such, we use the condition of human answerability for an annotated question-answer pair as follows:
It is answerable if at least one of three additional non-expert human validators can provide an answer matching the original. 
We conduct answerability checks on both the validation and test sets, and achieve answerability scores of $87.95\%$, $85.41\%$, and $82.63\%$ for \dataset{BiDAF}, \dataset{BERT}, and \dataset{RoBERTa}. 
We discard all questions deemed unanswerable from the validation and test sets, and further discard all data from any workers with less than half of their questions considered answerable. 
It should be emphasised that the main purpose of this process is to create a level playing field for comparison across datasets constructed for different model adversaries, and can inevitably result in valid questions being discarded.
The total cost for training and qualification, dataset construction, and validation is approximately USD 27,000.
}

\paragraph{Human Performance}{
We select a randomly chosen validator's answer to each question and compute Exact Match (EM) and word overlap F$_\text{1}$ scores with the original to calculate non-expert human performance; Table~\ref{tab:human_performance} shows the result.
We observe a clear trend:
the stronger the model in the loop used to construct the dataset, the harder the resulting questions become for humans.
}

%%%%%%%%%%%%%%%%%%%%%%%%%%%%%%%%%%%%%%%%%%%%%%%%
\begin{table}[t]
    \aboverulesep=0pt
    \belowrulesep=0pt
    \renewcommand{\arraystretch}{1.2}
    \centering
    \setlength{\tabcolsep}{10.6pt}
        \begin{tabular} {l ac ac@{}}
                 & \multicolumn{2}{c}{\emph{Dev}} & \multicolumn{2}{c}{\emph{Test}} \\ 
                 \hhline{~--||--}
            \textbf{Resource}  &\emph{EM}&\emph{F$_\text{1}$}&\emph{EM}&\emph{F$_\text{1}$} \\
            \toprule
            \dataset{BiDAF}    &  63.0  &  76.9  & 62.6  & 78.5   \\
            \dataset{BERT}     &  59.2  &  74.3  & 63.9  & 76.9   \\
            \dataset{RoBERTa}  &  58.1  &  72.0  & 58.7  & 73.7   \\
        \bottomrule
        \end{tabular}
    \caption{Non-expert human performance results for a randomly-selected validator per question.}
    \label{tab:human_performance}
\end{table}
%%%%%%%%%%%%%%%%%%%%%%%%%%%%%%%%%%%%%%%%%%%%%%%%

\subsection{Dataset Statistics}

Table~\ref{tab:statistics_sizes} provides general details on the number of passages and question-answer pairs used in the different dataset splits.
The average number of words in questions and answers, as well as the average longest n-gram overlap between passage and question are given in Table~\ref{tab:statistics_compare}.
%
%%%%%%%%%%%%%%%%%%%%%%%%%%%%%%%%%%%%%%%%%%%%%%%%
\begin{table}[t]
    \centering
    \footnotesize
    \setlength{\tabcolsep}{4pt}
        \begin{tabular} {@{\extracolsep{1pt}}lrrrrrr@{}}
                 & \multicolumn{3}{c}{\emph{\#Passages}} & \multicolumn{3}{c}{\emph{\#QAs}} \\ 
                 \cline{2-4} \cline{5-7}
            \textbf{Resource}  &\emph{Train}&\emph{Dev}&\emph{Test}&\emph{Train}&\emph{Dev}&\emph{Test} \\
            \toprule
            \dataset{SQuAD}    &  18,891  &  971  & 1,096  & 87,599 & 5,278 & 5,292 \\
            \dataset{BiDAF}    &  2,523  &  278  & 277  & 10,000 & 1,000 & 1,000   \\
            \dataset{BERT}     &  2,444  &  283  & 292  & 10,000 & 1,000 & 1,000 \\
            \dataset{RoBERTa}  &  2,552  &  341  & 333  & 10,000 & 1,000 & 1,000 \\
        \bottomrule
        \end{tabular}
    \caption{Number of passages and question-answer pairs for each data resource.}
    \label{tab:statistics_sizes}
\end{table}
%%%%%%%%%%%%%%%%%%%%%%%%%%%%%%%%%%%%%%%%%%%%%%%%

\begin{figure}[t]
    \centering
    \includegraphics[width=\columnwidth]{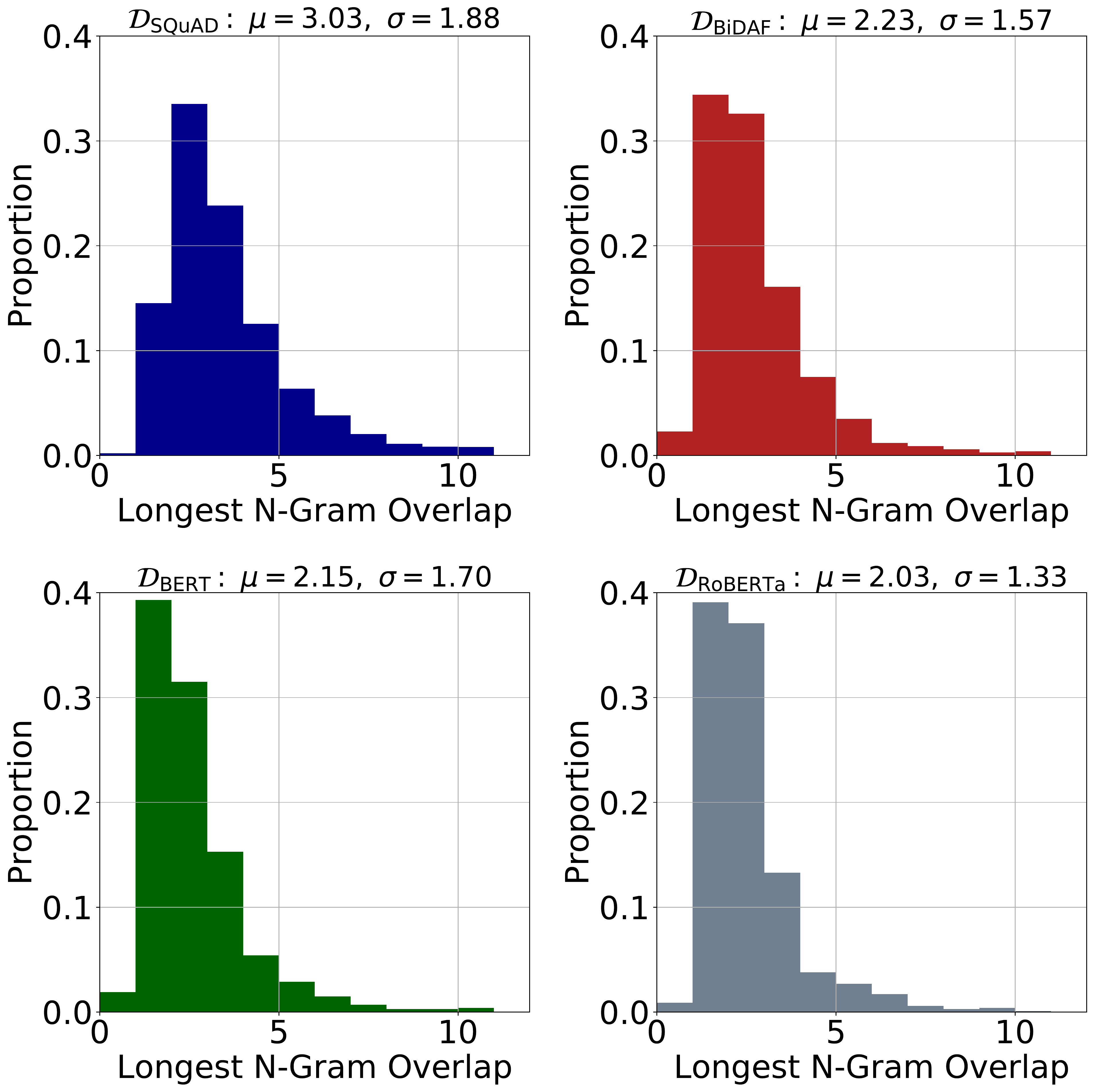}
    \caption{Distribution of longest n-gram overlap between passage and question for different datasets. $\mu$:~mean; $\sigma$: standard deviation.} 
    \label{fig:hists_ngram_overlap}
\end{figure}

%
We can again observe two clear trends: From weaker towards stronger models used in the annotation loop, the average length of answers increases, and the largest n-gram overlap drops from 3 to 2 tokens.
That is, on average there is a trigram overlap between the passage and question for \dataset{SQuAD}, but only a bigram overlap for \dataset{RoBERTa} (Figure~\ref{fig:hists_ngram_overlap}).\footnote{Note that the original \squadone{} dataset can be considered a limit case of the adversarial annotation framework, in which the model in the loop always predicts the wrong answer, thus every question is accepted.}
This is in line with prior observations on lexical overlap as a predictive cue in \squad~\cite{weissenborn-etal-2017-making,min2018efficient}; questions with less overlap are harder to answer for any of the three models. 
%

%
%%%%%%%%%%%%%%%%%%%%%%%%%%%%%%
%%%   FIGURE  %%%%
%%%%%%%%%%%%%%%%%%%%%%%%%%%%%%
%%%%%%%%%%%%%%%%%%%%%%%%%%%%%%

%%%%%%%%%%%%%%%%%%%%%%%%%%%%%%%%%%%%%%%%%%%%%%%%
\begin{table}[t]
    \centering
    \footnotesize
    \setlength{\tabcolsep}{3pt}
        \begin{tabular} {@{\extracolsep{1pt}}lrrrr@{}}
                 & \dataset{SQuAD} & \dataset{BiDAF} & \dataset{BERT} & \dataset{RoBERTa} \\ 
            \toprule
            % Passage length    &  118.1  &  115.3  & 114.6  & 114.6   \\
            Question length     &  10.3  &  9.8  & 9.8  & 10.0   \\
            Answer length  &  2.6  &  2.9  & 3.0  & 3.2   \\
            N-Gram overlap  &  3.0  &  2.2  & 2.1  & 2.0   \\
        \bottomrule
        \end{tabular}
    \caption{Average number of words per question and answer, and average longest n-gram overlap between passage and question.}
    \label{tab:statistics_compare}
\end{table}
%%%%%%%%%%%%%%%%%%%%%%%%%%%%%%%%%%%%%%%%%%%%%%%%

We furthermore analyse question types based on the question \textit{wh-}word.
We find that -- in contrast to \dataset{SQuAD} -- the datasets collected with a model in the annotation loop have fewer \textit{when}, \textit{how} and \textit{in} questions, and more \textit{which}, \textit{where} and \textit{why} questions, as well as questions in the \textit{other} category, which indicates increased question diversity.
In terms of answer types, we observe more common noun and verb phrase clauses than in \dataset{SQuAD}, as well as fewer dates, names, and numeric answers.
This reflects on the strong answer-type matching capabilities of contemporary RC models.
\ifarxiv For further dataset statistics on this, see Appendix~\ref{sec:appendix_dataset_statistics}.\fi
The training and validation sets used in this analysis (\dataset{BiDAF}, \dataset{BERT} and \dataset{RoBERTa}) will be publicly released.
% {\ifarxiv}{\else} upon acceptance{\fi}. 
%

%

%%%%%%%%%%%%%%%%%%%%%%%%%%%%%%%%%%%%%%%%%%%%%%%%
\begin{table}[t]
    \aboverulesep=0pt
    \belowrulesep=0pt
    \renewcommand{\arraystretch}{1.2}
    \centering
    \footnotesize
    \setlength{\tabcolsep}{5.4pt}
        \begin{tabular} {@{\extracolsep{0pt}}llacac@{}}
                 && \multicolumn{2}{c}{\emph{Original}} & \multicolumn{2}{c}{\emph{Re-init.}} \\ 
                 \hhline{~~--||--}
            \textbf{Model} & \textbf{Resource}  &\emph{EM}&\emph{F$_\text{1}$}&\emph{EM}&\emph{F$_\text{1}$} \\
            \toprule
            BiDAF   &   \datasetsplit{BiDAF}{dev}    &  0.0  &  5.3  & 10.7\std{0.8}  & 20.4\std{1.0}   \\
            BERT    &   \datasetsplit{BERT}{dev}     &  0.0  &  4.9  & 19.7\std{1.0}  & 30.1\std{1.2}   \\
            RoBERTa &   \datasetsplit{RoBERTa}{dev}  &  0.0  &  6.1  & 15.7\std{0.9}  & 25.8\std{1.2}   \\
            \midrule
            BiDAF   &   \datasetsplit{BiDAF}{test}    &  0.0  & 5.5  & 11.6\std{1.0}  & 21.3\std{1.2}   \\
            BERT    &   \datasetsplit{BERT}{test}     &  0.0  & 5.3  & 18.9\std{1.2}  & 29.4\std{1.1}   \\
            RoBERTa &   \datasetsplit{RoBERTa}{test}  &  0.0  & 5.9  & 16.1\std{0.8}  & 26.7\std{0.9}   \\
        \bottomrule
        \end{tabular}
    \caption{Consistency of the adversarial effect (or lack thereof) when retraining the models in the loop on the same data again, but with different random seeds. We report the mean and standard deviation (subscript) over 10 re-initialisation runs.}
    \label{tab:replication_results}
\end{table}
%%%%%%%%%%%%%%%%%%%%%%%%%%%%%%%%%%%%%%%%%%%%%%%%

\section{Experiments}

\subsection{Consistency of the Model in the Loop}
We begin with an experiment regarding the consistency of the adversarial nature of the models in the annotation loop.
Our annotation pipeline is designed to reject all samples where the model correctly predicts the answer.
How reproducible is this when retraining the model with the same training data?
To measure this, we evaluate the performance of instances of BiDAF, BERT, and RoBERTa, which only differ from the model used during annotation in their random initialisation and order of mini-batch samples during training.
These results are shown in Table~\ref{tab:replication_results}.

First, we observe -- as expected given our annotation constraints -- that model performance is 0.0EM on datasets created with the same respective model in the annotation loop. 
We observe however that retrained models do not reliably perform as poorly on those samples.
For example, BERT reaches 19.7EM, whereas the original model used during annotation provides no correct answer with 0.0EM.
This demonstrates that random model components can substantially affect the adversarial annotation process.
The evaluation furthermore serves as a baseline for subsequent model evaluations: This much of the performance range can be learnt merely by retraining the same model.
A possible takeaway for employing the model-in-the-loop annotation strategy in the future is to rely on ensembles of adversaries and reduce the dependency on one particular model instantiation, as investigated by~\citet{DBLP:journals/corr/abs-1811-09300}.

%%%%%%%%%%%%%%%%%%%%%%%%%%%%%%%%%%%%%%%%%%%%%%%%
\begin{table*}[t]
    \aboverulesep=0pt
    \belowrulesep=0pt
    \renewcommand{\arraystretch}{1.2}
    \centering
    \footnotesize
    \setlength{\tabcolsep}{2.7pt}
        \begin{tabular} {@{\extracolsep{0pt}}l | l | ac | ac ac ac | ac ac@{}}
                &&  \multicolumn{12}{c}{\textbf{Evaluation (Test) Dataset}}\\
                \cline{3-4} \cline{4-10} \cline{10-14}
                \textbf{Model} & \textbf{Trained On} & \multicolumn{2}{c|}{\textbf{\dataset{SQuAD}}} & \multicolumn{2}{c}{\textbf{\dataset{BiDAF}}} & \multicolumn{2}{c}{\textbf{\dataset{BERT}}} & \multicolumn{2}{c|}{\textbf{\dataset{RoBERTa}}} & \multicolumn{2}{c}{\textbf{\dataset{DROP}}} & \multicolumn{2}{c}{\textbf{\dataset{NQ}}} \\
                \hhline{~|~|--|--|--|--|--|--}
             &  &\emph{EM}&\emph{F$_\text{1}$}&\emph{EM}&\emph{F$_\text{1}$}&\emph{EM}&\emph{F$_\text{1}$} &\emph{EM}&\emph{F$_\text{1}$} &\emph{EM}&\emph{F$_\text{1}$} &\emph{EM}&\emph{F$_\text{1}$} \\
        
        \toprule
                            & \emph{\dataset{SQuAD(10K)}}   & \resultsemph{40.9}\std{0.6} & \resultsemph{54.3}\std{0.6} & 7.1\std{0.6} & \resultsemph{15.7}\std{0.6} & 5.6\std{0.3} & 13.5\std{0.4} & 5.7\std{0.4} & 13.5\std{0.4} & 3.8\std{0.4} & 8.6\std{0.6} & \resultsemph{25.1}\std{1.1} & \resultsemph{38.7}\std{0.7} \\
            \emph{BiDAF}    & \emph{\dataset{BiDAF}}        & 11.5\std{0.4} & 20.9\std{0.4} & 5.3\std{0.4} & 11.6\std{0.5} & 7.1\std{0.4} & 14.8\std{0.6} & 6.8\std{0.5} & 13.5\std{0.6} & 6.5\std{0.5} & 12.4\std{0.4} & 15.7\std{1.1} & 28.7\std{0.8} \\
                            & \emph{\dataset{BERT}}         &  10.8\std{0.3} & 19.8\std{0.4} & \resultsemph{7.2}\std{0.5} & 14.4\std{0.6} & 6.9\std{0.3} & 14.5\std{0.4} & 8.1\std{0.4} & 15.0\std{0.6} & 7.8\std{0.9} & 14.5\std{0.9} & 16.5\std{0.6} & 28.3\std{0.9} \\
                            & \emph{\dataset{RoBERTa}}      & 10.7\std{0.2} & 20.2\std{0.3} & 6.3\std{0.7} & 13.5\std{0.8} & \resultsemph{9.4}\std{0.6} & \resultsemph{17.0}\std{0.6} & \resultsemph{8.9}\std{0.9} & \resultsemph{16.0}\std{0.8} & \resultsemph{15.3}\std{0.8} & \resultsemph{22.9}\std{0.8} & 13.4\std{0.9} & 27.1\std{1.2} \\
        
        \midrule
        
                            & \emph{\dataset{SQuAD(10K)}}   & \resultsemph{69.4}\std{0.5} & \resultsemph{82.7}\std{0.4} & 35.1\std{1.9} & 49.3\std{2.2} & 15.6\std{2.0} & 27.3\std{2.1} & 11.9\std{1.5} & 23.0\std{1.4} & 18.9\std{2.3} & 28.9\std{3.2} & 52.9\std{1.0} & 68.2\std{1.0} \\
            \emph{BERT}     & \emph{\dataset{BiDAF}}        & 66.5\std{0.7} & 80.6\std{0.6} & \resultsemph{46.2}\std{1.2} & \resultsemph{61.1}\std{1.2} & \resultsemph{37.8}\std{1.4} & \resultsemph{48.8}\std{1.5} & \resultsemph{30.6}\std{0.8} & \resultsemph{42.5}\std{0.6} & \resultsemph{41.1}\std{2.3} & \resultsemph{50.6}\std{2.0} & \resultsemph{54.2}\std{1.2} & \resultsemph{69.8}\std{0.9} \\
                            & \emph{\dataset{BERT}}         & 61.2\std{1.8} & 75.7\std{1.6} & 42.9\std{1.9} & 57.5\std{1.8} & 37.4\std{2.1} & 47.9\std{2.0} & 29.3\std{2.1} & 40.0\std{2.3} & 39.4\std{2.2} & 47.6\std{2.2} & 49.9\std{2.3} & 65.7\std{2.3} \\
                            & \emph{\dataset{RoBERTa}}      & 57.0\std{1.7} & 71.7\std{1.8} & 37.0\std{2.3} & 52.0\std{2.5} & 34.8\std{1.5} & 45.9\std{2.0} & 30.5\std{2.2} & 41.2\std{2.2} & 39.0\std{3.1} & 47.4\std{2.8} & 45.8\std{2.4} & 62.4\std{2.5} \\
        
        \midrule
        
                            & \emph{\dataset{SQuAD(10K)}}   & \resultsemph{68.6}\std{0.5} & \resultsemph{82.8}\std{0.3} & 37.7\std{1.1} & 53.8\std{1.1} & 20.8\std{1.2} & 34.0\std{1.0} & 11.0\std{0.8} & 22.1\std{0.9} & 25.0\std{2.2} & 39.4\std{2.4} & 43.9\std{3.8} & 62.8\std{3.1} \\ 
            \emph{RoBERTa}  & \emph{\dataset{BiDAF}}        & 64.8\std{0.7} & 80.0\std{0.4} & \resultsemph{48.0}\std{1.2} & \resultsemph{64.3}\std{1.1} & \resultsemph{40.0}\std{1.5} & \resultsemph{51.5}\std{1.3} & 29.0\std{1.9} & 39.9\std{1.8} & \resultsemph{44.5}\std{2.1} & \resultsemph{55.4}\std{1.9} & \resultsemph{48.4}\std{1.1} & \resultsemph{66.9}\std{0.8} \\
                            & \emph{\dataset{BERT}}         & 59.5\std{1.0} & 75.1\std{0.9} & 45.4\std{1.5} & 60.7\std{1.5} & 38.4\std{1.8} & 49.8\std{1.7} & 28.2\std{1.5} & 38.8\std{1.5} & 42.2\std{2.3} & 52.6\std{2.0} & 45.8\std{1.1} & 63.6\std{1.1} \\
                            & \emph{\dataset{RoBERTa}}      & 56.2\std{0.7} & 72.1\std{0.7} & 41.4\std{0.8} & 57.1\std{0.8} & 38.4\std{1.1} & 49.5\std{0.9} & \resultsemph{30.2}\std{1.3} & \resultsemph{41.0}\std{1.2} & 41.2\std{0.9} & 51.2\std{0.8} & 43.6\std{1.1} & 61.6\std{0.9} \\
        \bottomrule
        \end{tabular}
    \caption{Training models on various datasets, each with 10,000 samples, and measuring their generalisation to different evaluation datasets. Results \underline{underlined} indicate the best result per model. We report the mean and standard deviation (subscript) over 10 runs with different random seeds.}
    \label{tab:train_only_on_adversarial_data}
\end{table*}
%%%%%%%%%%%%%%%%%%%%%%%%%%%%%%%%%%%%%%%%%%%%%%%%

\subsection{Adversarial Generalisation}
%
% Key results to highlight:
% Table 6:
% - 1. Negative performance progression when evaluated against datasets constructed with a stronger model in the loop
% - 2. BiDAF cannot generalise well to datasets constructed with a model in the loop
% - 3. BERT and RoBERTa are able to partially overcome their blind spots through adversarial training
% - 4. Performance on SQuAD decreases for the same model when trained on data collected with progressively stronger models
% - 5. Performance improvements on DROP when trained with adversarial datasets, also reasonable performance on NQs which decreases with model hardness – NOTE how using D_BiDAF is best
%
% Table 7:
% - 1. Similar effects with BiDAF struggling to learn much about the adversarially constructed datasets
% - 2. 0 EM for datasets collected with the same model-in-the-loop
% - 3. BiDAF can’t learn much, but it can answer some of the questions in D(BERT) and D(RoBERTa)
% - 4.  Performance on SQuAD tends to improve slightly be adding adversarially constructed data (also for RoBERTa)
% - 5. RoBERTa performance improves on D(RoBERTa) with increasing model hardness
%
%
A potential problem with the focus on challenging questions is that they might be very distinct from one another, leading to difficulties in learning to generalise to and from them.
We conduct a series of experiments in which we train on \dataset{BiDAF}, \dataset{BERT}, and \dataset{RoBERTa}, and observe how well models can learn to generalise to the respective test portions of these datasets.
Table~\ref{tab:train_only_on_adversarial_data} shows the results, and there is a multitude of observations.

First, one clear trend we observe across all training data setups is a negative performance progression when evaluated against datasets constructed with a stronger model in the loop.
This trend holds true for all but the BiDAF model, in each of the training configurations, and for each of the evaluation datasets. 
For example, RoBERTa trained on \dataset{RoBERTa} achieves 72.1, 57.1, 49.5, and 41.0F$_\text{1}$ when evaluated on \dataset{SQuAD}, \dataset{BiDAF}, \dataset{BERT}, and \dataset{RoBERTa} respectively.
Second, we observe that the BiDAF model is not able to generalise well to datasets constructed with a model in the loop, independent of its training setup.
In particular it is unable to learn from \dataset{BiDAF}, thus failing to overcome some of its own blind spots through adversarial training.
Irrespective of the training dataset, BiDAF consistently performs poorly on the adversarially collected evaluation datasets, and we also note a substantial performance drop when trained on \dataset{BiDAF}, \dataset{BERT}, or \dataset{RoBERTa} and evaluated on \dataset{SQuAD}.
In contrast, BERT and RoBERTa are able to partially overcome their blind spots through training on data collected with a model in the loop, and to a degree that far exceeds what would be expected from random retraining~(cf.~Table~\ref{tab:replication_results}).
For example, BERT reaches 47.9F$_\text{1}$ when trained and evaluated on \dataset{BERT}, while RoBERTa trained on \dataset{RoBERTa} reaches 41.0F$_\text{1}$ on \dataset{RoBERTa}, both considerably better than random retraining, or when training on the non-adversarially collected \dataset{SQuAD(10K)} showing gains of 20.6F$_\text{1}$ for BERT and 18.9F$_\text{1}$ for RoBERTa.
These observations suggest that there exists learnable structure among harder questions that can be picked up by some of the models, yet not all, as BiDAF fails to achieve this.
The fact that even BERT can learn to generalise to \dataset{RoBERTa}, but not BiDAF to \dataset{BERT} suggests the existence of an inherent limitation to what BiDAF can learn from these new samples, compared to BERT and RoBERTa.
More generally, we observe that training on \dataset{S}, where $S$ is a stronger RC model, helps generalise to \dataset{W}, where $W$ is a weaker model, for example, training on \dataset{RoBERTa} and testing on \dataset{BERT}.
On the other hand, training on \dataset{W} also leads to generalisation towards \dataset{S}. 
For example, RoBERTa trained on 10,000 \squad{} samples reaches 22.1F$_\text{1}$ on \dataset{RoBERTa} (\dataset{S}), whereas training RoBERTa on \dataset{BiDAF} and \dataset{BERT} (\dataset{W}) bumps this number to 39.9F$_\text{1}$ and 38.8F$_\text{1}$, respectively.
Third, we observe similar performance degradation patterns for both BERT and RoBERTa on \dataset{SQuAD} when trained on data collected with increasingly stronger models in the loop. For example, RoBERTa evaluated on \dataset{SQuAD} achieves 82.8, 80.0, 75.1, and 72.1F$_\text{1}$ when trained on \dataset{SQuAD(10K)}, \dataset{BiDAF}, \dataset{BERT}, and \dataset{RoBERTa} respectively. This may indicate a gradual shift in the distributions of composed questions as the model in the loop gets stronger.
These observations suggest an encouraging takeaway for the model-in-the-loop annotation paradigm: 
Even though a particular model might be chosen as an adversary in the annotation loop, which at some point falls behind more recent state-of-the-art models, these future models can still benefit from data collected with the weaker model, and also generalise better to samples composed with the stronger model in the loop.
%

%%%%%%%%%%%%%%%%%%%%%%%%%%%%%%%%%%%%%%%%%%%%%%%%
\begin{table*}[t]
    \aboverulesep=0pt
    \belowrulesep=0pt
    \renewcommand{\arraystretch}{1.2}
    \centering
    \setlength{\tabcolsep}{3.6pt}
        \begin{tabular} {@{\extracolsep{0pt}}l | l | ac ac ac ac@{}}
                &&  \multicolumn{8}{c}{\textbf{Evaluation (Test) Dataset}}\\
                \cline{3-10}
                \textbf{Model} & \textbf{Training Dataset} & \multicolumn{2}{c}{\dataset{SQuAD}} & \multicolumn{2}{c}{\textbf{\dataset{BiDAF}}} & \multicolumn{2}{c}{\textbf{\dataset{BERT}}} &\multicolumn{2}{c}{\textbf{\dataset{RoBERTa}}} \\
                \hhline{~|~|--||--||--||--}
             &  &\emph{EM}&\emph{F$_\text{1}$}&\emph{EM}&\emph{F$_\text{1}$}&\emph{EM}&\emph{F$_\text{1}$} &\emph{EM}&\emph{F$_\text{1}$} \\
             
        \toprule
        
                                & \emph{\dataset{SQuAD}}  & \resultsemph{56.7}\std{0.5} & \resultsemph{70.1}\std{0.3} & 11.6\std{1.0} & 21.3\std{1.1} & 8.6\std{0.6} & 17.3\std{0.8} & 8.3\std{0.7} & 16.8\std{0.5} \\
            \emph{BiDAF}        & \emph{\dataset{SQuAD} + \dataset{BiDAF}}  & 56.3\std{0.6} & 69.7\std{0.4} & 14.4\std{0.9} & 24.4\std{0.9} & 15.6\std{1.1} & 24.7\std{1.1} & 14.3\std{0.5} & 23.3\std{0.7} \\
                                & \emph{\dataset{SQuAD} + \dataset{BERT}}  & 56.2\std{0.6} & 69.4\std{0.6} & 14.4\std{0.7} & 24.2\std{0.8} & 15.7\std{0.6} & 25.1\std{0.6} & 13.9\std{0.8} & 22.7\std{0.8} \\
                                & \emph{\dataset{SQuAD} + \dataset{RoBERTa}}  & 56.2\std{0.7} & 69.6\std{0.6} & \resultsemph{14.7}\std{0.9} & \resultsemph{24.8}\std{0.8} & \resultsemph{17.9}\std{0.5} & \resultsemph{26.7}\std{0.6} & \resultsemph{16.7}\std{1.1} & \resultsemph{25.0}\std{0.8} \\

        \midrule
        
                                & \emph{\dataset{SQuAD}}  & 74.8\std{0.3} & 86.9\std{0.2} & 46.4\std{0.7} & 60.5\std{0.8} & 24.4\std{1.2} & 35.9\std{1.1} & 17.3\std{0.7} & 28.9\std{0.9} \\
            \emph{BERT}         & \emph{\dataset{SQuAD} + \dataset{BiDAF}}  & 75.2\std{0.4} & \resultsemph{87.2}\std{0.2} & 52.4\std{0.9} & 66.5\std{0.9} & 40.9\std{1.3} & 51.2\std{1.5} & 32.9\std{0.9} & 44.1\std{0.8} \\
                                & \emph{\dataset{SQuAD} + \dataset{BERT}}  & 75.1\std{0.3} & 87.1\std{0.3} & \resultsemph{54.1}\std{1.0} & \resultsemph{68.0}\std{0.8} & 43.7\std{1.1} & 54.1\std{1.3} & 34.7\std{0.7} & 45.7\std{0.8} \\
                                & \emph{\dataset{SQuAD} + \dataset{RoBERTa}}  & \resultsemph{75.3}\std{0.4} & 87.1\std{0.3} & 53.0\std{1.1} & 67.1\std{0.8} & \resultsemph{44.1}\std{1.1} & \resultsemph{54.4}\std{0.9} & \resultsemph{36.6}\std{0.8} & \resultsemph{47.8}\std{0.5} \\

        \midrule
        
                                & \emph{\dataset{SQuAD}}  & 73.2\std{0.4} & 86.3\std{0.2} & 48.9\std{1.1} & 64.3\std{1.1} & 31.3\std{1.1} & 43.5\std{1.2} & 16.1\std{0.8} & 26.7\std{0.9} \\
            \emph{RoBERTa}      & \emph{\dataset{SQuAD} + \dataset{BiDAF}}  & \resultsemph{73.9}\std{0.4} & \resultsemph{86.7}\std{0.2} & 55.0\std{1.4} & 69.7\std{0.9} & 46.5\std{1.1} & 57.3\std{1.1} & 31.9\std{0.8} & 42.4\std{1.0} \\
                                & \emph{\dataset{SQuAD} + \dataset{BERT}}  & 73.8\std{0.2} & \resultsemph{86.7}\std{0.2} & 55.4\std{1.0} & 70.1\std{0.9} & 48.9\std{1.0} & 59.0\std{1.2} & 32.9\std{1.3} & 43.7\std{1.4} \\
                                & \emph{\dataset{SQuAD} + \dataset{RoBERTa}}  & 73.5\std{0.3} & 86.5\std{0.2} & \resultsemph{55.9}\std{0.7} & \resultsemph{70.6}\std{0.7} & \resultsemph{49.1}\std{1.2} & \resultsemph{59.5}\std{1.2} & \resultsemph{34.7}\std{1.0} & \resultsemph{45.9}\std{1.2} \\
        \bottomrule
        \end{tabular}
    \caption{Training models on \squad{}, as well as \squad{} combined with different adversarially created datasets. 
    \ifarxiv Results \underline{underlined} indicate the best result per model.\fi
    We report the mean and standard deviation (subscript) over 10 runs with different random seeds.}
    \label{tab:train_on_squad_and_adversarial_data}
\end{table*}
%%%%%%%%%%%%%%%%%%%%%%%%%%%%%%%%%%%%%%%%%%%%%%%%
%

%
We further show experimental results for the same models and training datasets, but now including \squad{} as additional training data in Table~\ref{tab:train_on_squad_and_adversarial_data}.
In this training setup we generally see improved generalisation to \dataset{BiDAF}, \dataset{BERT}, and \dataset{RoBERTa}.
Interestingly, the relative differences between \dataset{BiDAF}, \dataset{BERT}, and \dataset{RoBERTa} as training sets used in conjunction with \squad{} are much diminished, and especially \dataset{RoBERTa} as (part of) the training set now generalises substantially better.
We see that BERT and RoBERTa both show consistent performance gains with the addition of the original \squadone{} training data, but unlike in Table~\ref{tab:train_only_on_adversarial_data}, this comes without any noticeable decline in performance on \dataset{SQuAD}, suggesting that the adversarially constructed datasets expose inherent model weaknesses, as investigated by~\citet{liu-etal-2019-inoculation}.
Furthermore, RoBERTa achieves the strongest results on the adversarially collected evaluation sets, in particular when trained on \dataset{SQuAD}~+~\dataset{RoBERTa}.
This stands in contrast to the results in Table~\ref{tab:train_only_on_adversarial_data}, where training on \dataset{BiDAF} in several cases led to better generalisation than training on \dataset{RoBERTa}.
A possible explanation is that training on \dataset{RoBERTa} leads to a larger degree of overfitting to specific adversarial examples in \dataset{RoBERTa} than training on \dataset{BiDAF}, and that the inclusion of a large number of standard \squad{} training samples can mitigate this effect.
Results for the models trained on all the datasets combined (\dataset{SQuAD}, \dataset{BiDAF}, \dataset{BERT}, and \dataset{RoBERTa}) are shown in Table~\ref{tab:train_on_squad_and_all_adversarial_data}.
These further support the previous observations and provide additional performance gains where, for example, RoBERTa achieves F$_\text{1}$ scores of 86.9 on \dataset{SQuAD}, 74.1 on \dataset{BiDAF}, 65.1 on \dataset{BERT}, and 52.7 on \dataset{RoBERTa}, surpassing the best previous performance on all adversarial datasets.
Finally, we identify a risk of datasets constructed with weaker models in the loop becoming outdated. 
For example, RoBERTa achieves 58.2EM/73.2F$_\text{1}$ on \dataset{BiDAF}, in contrast to 0.0EM/5.5F$_\text{1}$ for BiDAF -- which is not far from the non-expert human performance of 62.6EM/78.5F$_\text{1}$ (cf.~Table \ref{tab:human_performance}).
It is also interesting to note that, even when training on all the combined data (cf. Table~\ref{tab:train_on_squad_and_all_adversarial_data}), BERT outperforms RoBERTa on \dataset{RoBERTa} and vice versa, suggesting that there may exist weaknesses inherent to each model class.

%%%%%%%%%%%%%%%%%%%%%%%%%%%%%%%%%%%%%%%%%%%%%%%%
\begin{table*}[t]
    \aboverulesep=0pt
    \belowrulesep=0pt
    \renewcommand{\arraystretch}{1.2}
    \centering
    \setlength{\tabcolsep}{9.6pt}
        \begin{tabular} {l | ac ac ac ac@{}}
                &  \multicolumn{8}{c}{\textbf{Evaluation (Test) Dataset}}\\
                \cline{2-9}
                \textbf{Model} & \multicolumn{2}{c}{\dataset{SQuAD}} & \multicolumn{2}{c}{\textbf{\dataset{BiDAF}}} & \multicolumn{2}{c}{\textbf{\dataset{BERT}}} &\multicolumn{2}{c}{\textbf{\dataset{RoBERTa}}} \\
                \hhline{~|--||--||--||--}
             & \emph{EM}&\emph{F$_\text{1}$}&\emph{EM}&\emph{F$_\text{1}$}&\emph{EM}&\emph{F$_\text{1}$} &\emph{EM}&\emph{F$_\text{1}$} \\
             
        \toprule
        
            \emph{BiDAF}        & 57.1\std{0.4} & 70.4\std{0.3} & 17.1\std{0.8} & 27.0\std{0.9} & 20.0\std{1.0} & 29.2\std{0.8} & 18.3\std{0.6} & 27.4\std{0.7} \\
            \emph{BERT}         & \resultsemph{75.5}\std{0.2} & \resultsemph{87.2}\std{0.2} & 57.7\std{1.0} & 71.0\std{1.1} & 52.1\std{0.7} & 62.2\std{0.7} & \resultsemph{43.0}\std{1.1} & \resultsemph{54.2}\std{1.0} \\
            \emph{RoBERTa}         & 74.2\std{0.3} & 86.9\std{0.3} & \resultsemph{59.8}\std{0.5} & \resultsemph{74.1}\std{0.6} & \resultsemph{55.1}\std{0.6} & \resultsemph{65.1}\std{0.7} & 41.6\std{1.0} & 52.7\std{1.0} \\

        \bottomrule
        \end{tabular}
    \caption{Training models on \squad{} combined with all the adversarially created datasets \dataset{BiDAF}, \dataset{BERT}, and \dataset{RoBERTa}. \ifarxiv Results \underline{underlined} indicate the best result per model.\fi
    We report the mean and standard deviation (subscript) over 10 runs with different random seeds.}
    \label{tab:train_on_squad_and_all_adversarial_data}
\end{table*}
%%%%%%%%%%%%%%%%%%%%%%%%%%%%%%%%%%%%%%%%%%%%%%%%

%%%%%%%%%%%%%%%%%%%%%%%%%%%%%%%%%%%%%%%%%%%%%%%%%%%%%%%%%%%%%%%%%%%%%%%%%%%%%%%%%%%%%%%%%%%%%%%%%%%%%%%%%%%%%%%%%%%%%%%%%%
\subsection{Generalisation to Non-Adversarial Data}
Compared to standard annotation, the model-in-the-loop approach generally results in new question distributions.
Consequently, models trained on adversarially composed questions might not be able to generalise to standard (``easy'') questions, thus limiting the practical usefulness of the resulting data.
To what extent do models trained on model-in-the-loop questions generalise differently to standard (``easy'') questions, compared to models trained on standard (``easy'') questions?
\ifpagebreaks
    \pagebreak
\fi
To measure this we further train each of our three models on either \dataset{BiDAF}, \dataset{BERT}, or \dataset{RoBERTa} and test on \dataset{SQuAD}, with results in the \dataset{SQuAD} columns of Table~\ref{tab:train_only_on_adversarial_data}.
For comparison, the models are also trained on 10,000 \squadone{} samples (referred to as \dataset{SQuAD(10K)}) chosen from the same passages as the adversarial datasets, thus eliminating size and paragraph choice as potential confounding factors.
The models are tuned for EM on the held-out \dataset{SQuAD} validation set. 
Note that, although performance values on the majority vote \dataset{SQuAD} dataset are lower than on the original, for the reasons described earlier, this enables direct comparisons across all datasets.

Remarkably, neither BERT nor RoBERTa show substantial drops when trained on \dataset{BiDAF} compared to training on \squad{} data ($-$2.1F$_\text{1}$, and $-$2.8F$_\text{1}$):
Training these models on a dataset with a weaker model in the loop still leads to strong generalisation even to data from the original \squad{} distribution, which all models in the loop are trained on.
BiDAF, on the other hand, fails to learn such information from the adversarially collected data, and drops >30F$_\text{1}$ for each of the new training sets, compared to training on \squad{}.

We also observe a gradual decrease in generalisation to \squad{} when training on \dataset{BiDAF} towards training on \dataset{RoBERTa}.
This suggests that the stronger the model, the more dissimilar the resulting data distribution becomes from the original \squad{} distribution.
We later find further support for this explanation in a qualitative analysis (Section~\ref{sec:qualitative}).
It may however also be due to a limitation of BERT and RoBERTa -- similar to BiDAF -- in learning from a data distribution designed to beat these models; an even stronger model might learn more from, for example, \dataset{RoBERTa}.

\subsection{Generalisation to DROP and NaturalQuestions}
Finally, we investigate to what extent models can transfer skills learned on the datasets created with a model in the loop to two recently introduced datasets: \drop{}~\cite{dua2019drop}, and \naturalquestions{}~\cite{kwiatkowski-etal-2019-natural}.
In this experiment we select the subsets of \drop{} and \naturalquestions{} that align with the structural constraints of \squad{} to ensure a like-for-like analysis.
Specifically, we only consider questions in \drop{} where the answer is a span in the passage and where there is only one candidate answer. 
For \naturalquestions{}, we consider all non-tabular long answers as passages, remove HTML tags and use the short answer as the extracted span. 
We apply this filtering on the validation sets for both datasets.
Next we split them, stratifying by document (as we did for \dataset{\squad{}}), which results in $1409 / 1418$ validation and test set examples for \drop{}, and $964 / 982$ for \naturalquestions{}, respectively.
We denote these datasets as \dataset{DROP} and \dataset{NQ} for clarity and distinction from their unfiltered versions. 
We consider the same models and training datasets as before, but tune on the respective validation sets of \dataset{DROP} and \dataset{NQ}.
Table~\ref{tab:train_only_on_adversarial_data} shows the results of these experiments in the respective \dataset{DROP} and \dataset{NQ} columns.

\begin{figure*}[t]
\includegraphics[width=\textwidth]{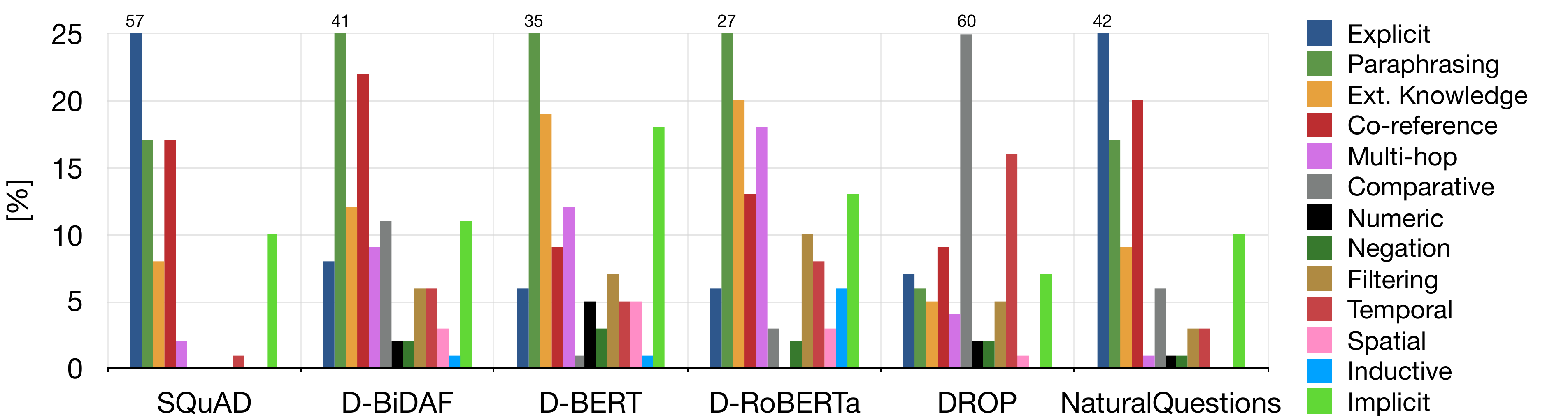}
\caption{Comparison of comprehension types for the questions in different datasets. The label types are neither mutually exclusive nor comprehensive. Values above columns indicate excess of the axis range.}\label{fig:reading_types}
\end{figure*}

First, we observe clear generalisation improvements towards \dataset{DROP} across all models compared to training on \dataset{SQuAD(10K)} when training on any of \dataset{BiDAF}, \dataset{BERT}, or \dataset{RoBERTa}.
That is, including a model in the loop for the training dataset leads to improved transfer towards \dataset{DROP}.
Note that \drop{} also makes use of a BiDAF model in the loop during annotation; these results are in line with our prior observations when testing the same setups on \dataset{BiDAF}, \dataset{BERT} and \dataset{RoBERTa}, compared to training on \dataset{SQuAD(10K)}.
\ifpagebreaks
    \pagebreak
\fi
Second, we observe overall strong transfer results towards \dataset{NQ}, with up to 69.8F$_\text{1}$ for a BERT model trained on \dataset{BiDAF}.
Note that this result is similar to, and even slightly improves over model training with \squad{} data of the same size.
That is, relative to training on SQuAD data, training on adversarially collected data \dataset{BiDAF} does not impede generalisation to the \dataset{NQ} dataset, which was created without a model in the annotation loop.
We then however see a similar negative performance progression as observed before when testing on \dataset{SQuAD}: the stronger the model in the annotation loop of the training dataset, the lower the test accuracy on test data from a data distribution composed without a model in the loop.

\section{Qualitative Analysis} \label{sec:qualitative}
Having applied the general model-in-the-loop methodology on models of varying strength, we next perform a qualitative comparison of the nature of the resulting questions.
As reference points we also include the original \squad{} questions, as well as \drop{} and \naturalquestions{} in this comparison: These datasets are both constructed to overcome limitations in \squad{} and have subsets sufficiently similar to \squad{} to make an analysis possible.
Specifically, we seek to understand the qualitative differences in terms of reading comprehension challenges posed by the questions in each of these datasets.

\subsection{Comprehension Requirements}
There exists a variety of prior work that seeks to understand the types of knowledge, comprehension skills or types of reasoning required to answer questions based on text~\cite{rajpurkar2016squad,clark2018think,sugawara2020assessing,dua2019drop,dasigi-etal-2019-quoref}; we are however unaware of any commonly accepted formalism.
We take inspiration from these but develop our own taxonomy of comprehension requirements which suits the datasets analysed\ifarxiv, see Appendix~\ref{sec:appendix_reasoning_types} for a detailed breakdown and examples of our annotation catalogue\fi.
{\ifarxiv}{\else}
Our taxonomy contains 13 labels, most of which are commonly used in other work.
However, the following three deserve additional clarification:
i) \textit{explicit} -- for which the answer is stated nearly word-for-word in the passage as it is in the question,
ii) \textit{filtering} -- a set of answers is narrowed down to select one by some particular distinguishing feature, and
iii) \textit{implicit} -- the answer builds on information implied by the passage and does not otherwise require any of the other types of reasoning.
{\fi}

We annotate questions with labels from this catalogue in a manner that is not mutually exclusive, and neither fully comprehensive; the development of such a catalogue is itself very challenging.
Instead, we focus on capturing the most salient characteristics of each given question, and assign it up to three of the labels in our catalogue.
In total, we analyse 100 samples from the validation set of each of the datasets; Figure~\ref{fig:reading_types} shows the results.

\subsection{Observations}

An initial observation is that the majority (57\%) of answers to \squad{} questions are stated explicitly, without comprehension requirements beyond the literal level.
This number decreases substantially for any of the model-in-the-loop datasets derived from \squad{} (e.g.,~8\% for \dataset{BiDAF}) and also \dataset{DROP}, yet 42\% of questions in \dataset{NQ} share this property.
In contrast to \squad{}, the model-in-the-loop questions generally tend to involve more paraphrasing.
They also require more external knowledge, and multi-hop inference (beyond co-reference resolution) with an increasing trend for stronger models used in the annotation loop.
Model-in-the-loop questions further fan out into a variety of small, but non-negligible proportions of more specific types of inference required for comprehension, for example, spatial or temporal inference (both going beyond explicitly stated spatial or temporal information) -- \squad{} questions rarely require these at all.
Some of these more particular inference types are common features of the other two datasets, in particular \emph{comparative} questions for \drop{} (60\%) and to a small extent also \naturalquestions{}.
Interestingly, \dataset{BiDAF} possesses the largest number of comparison questions (11\%) among our model-in-the-loop datasets, whereas \dataset{BERT} and \dataset{RoBERTa} only possess 1\% and 3\%, respectively.
This offers an explanation for our previous observation in Table~\ref{tab:train_only_on_adversarial_data}, where BERT and RoBERTa perform better on \dataset{DROP} when trained on \dataset{BiDAF} rather than on \dataset{BERT} or \dataset{RoBERTa}.
It is likely that BiDAF as a model in the loop is worse than BERT and RoBERTa at \emph{comparative} questions, as evidenced by the results in Table~\ref{tab:train_only_on_adversarial_data} with BiDAF reaching 8.6F$_\text{1}$, BERT reaching 28.9F$_\text{1}$, and RoBERTa reaching 39.4F$_\text{1}$ on \dataset{DROP} (when trained on \dataset{SQuAD(10K)}).

The distribution of {\naturalquestions} contains elements of both the \squad{} and \dataset{BiDAF} distributions, which offers a potential explanation for the strong performance on \dataset{NQ} of models trained on \dataset{SQuAD(10K)} and \dataset{BiDAF}.
Finally, the gradually shifting distribution away from both \squad{} and \naturalquestions{} as the model-in-the-loop strength increases reflects our prior observations on the decreasing performance on \squad{} and \naturalquestions{} of models trained on datasets with progressively stronger models in the loop.

\section{Discussion and Conclusions}
We have investigated an RC annotation paradigm that requires a model in the loop to be ``beaten'' by an annotator.
Applying this approach with progressively stronger models in the loop (BiDAF, BERT, and RoBERTa), we produced three separate datasets.
Using these datasets, we investigated several questions regarding the annotation paradigm, in particular whether such datasets grow outdated as stronger models emerge, and their generalisation to standard (non-adversarially collected) questions. 
We found that stronger models can still learn from data collected with a weak adversary in the loop, and their generalisation improves even on datasets collected with a stronger adversary.
Models trained on data collected with a model in the loop further generalise well to non-adversarially collected data, both on \squad~and on \naturalquestions, yet we observe a gradual deterioration in performance with progressively stronger adversaries.

We see our work as a contribution towards the emerging paradigm of model-in-the-loop annotation.
While this paper has focused on RC, with {\squad} as the original dataset used to train model adversaries, we see no reason in principle why findings would not be similar for other tasks using the same annotation paradigm, when crowdsourcing challenging samples with a model in the loop.
We would expect the insights and benefits conveyed by model-in-the-loop annotation to be the greatest on mature datasets where models exceed human performance: Here the resulting data provides a magnifying glass on model performance, focused in particular on samples which models struggle on.
On the other hand, applying the method to datasets where performance has not yet plateaued would likely result in a more similar distribution to the original data, which is challenging to models a priori.
We hope that the series of experiments on replicability, observations on transfer between datasets collected using models of different strength, as well as our findings regarding generalisation to non-adversarially collected data, can support and inform future research and annotation efforts using this paradigm.

\section*{Acknowledgements}
The authors would like to thank Christopher Potts for his detailed and constructive feedback, and our reviewers.
This work was supported by
the European Union's Horizon 2020 research and innovation programme under grant agreement No.~875160
and
the UK Defence Science and Technology Laboratory~(Dstl) and Engineering and Physical Research Council~(EPSRC) under grant EP/R018693/1 as a part of the collaboration between US DOD, UK MOD,
and UK EPSRC under the Multidisciplinary University Research Initiative~(MURI).

\bibliography{main}
\bibliographystyle{acl_natbib}

\ifappendix
    \clearpage
    \appendix
    
    \section{Additional Dataset Statistics}
    \label{sec:appendix_dataset_statistics}
    
    \paragraph{Question statistics}{
    In Figure~\ref{fig:hists_question_len} we analyse question lengths across \squadone{} and compare them to questions constructed with different models in the annotation loop.
    While the mean of the distributions is similar, there is more question length variability when using a model in the loop. 
    We also perform analysis of question types by \textit{wh-} word as described earlier (see Figure~\ref{fig:question_types}).
    This is in further detail displayed using sunburst plots of the first three question tokens for \dataset{SQuAD} (cf.~Figure~\ref{fig:sunburst_squad}), \dataset{BiDAF} (cf.~Figure~\ref{fig:sunburst_bidaf}), \dataset{BERT} (cf.~Figure~\ref{fig:sunburst_bert}) and \dataset{RoBERTa} (cf.~Figure~\ref{fig:sunburst_roberta}). 
    We observe a general trend towards more diverse questions with increasing model-in-the-loop strength.
    }

    %%%%%%%%%%%%%%%%%%%%%%%%%%%%%%
    %%%   FIGURE  %%%%
    %%%%%%%%%%%%%%%%%%%%%%%%%%%%%%
    \begin{figure}[t]
        \centering
        \includegraphics[width=\columnwidth]{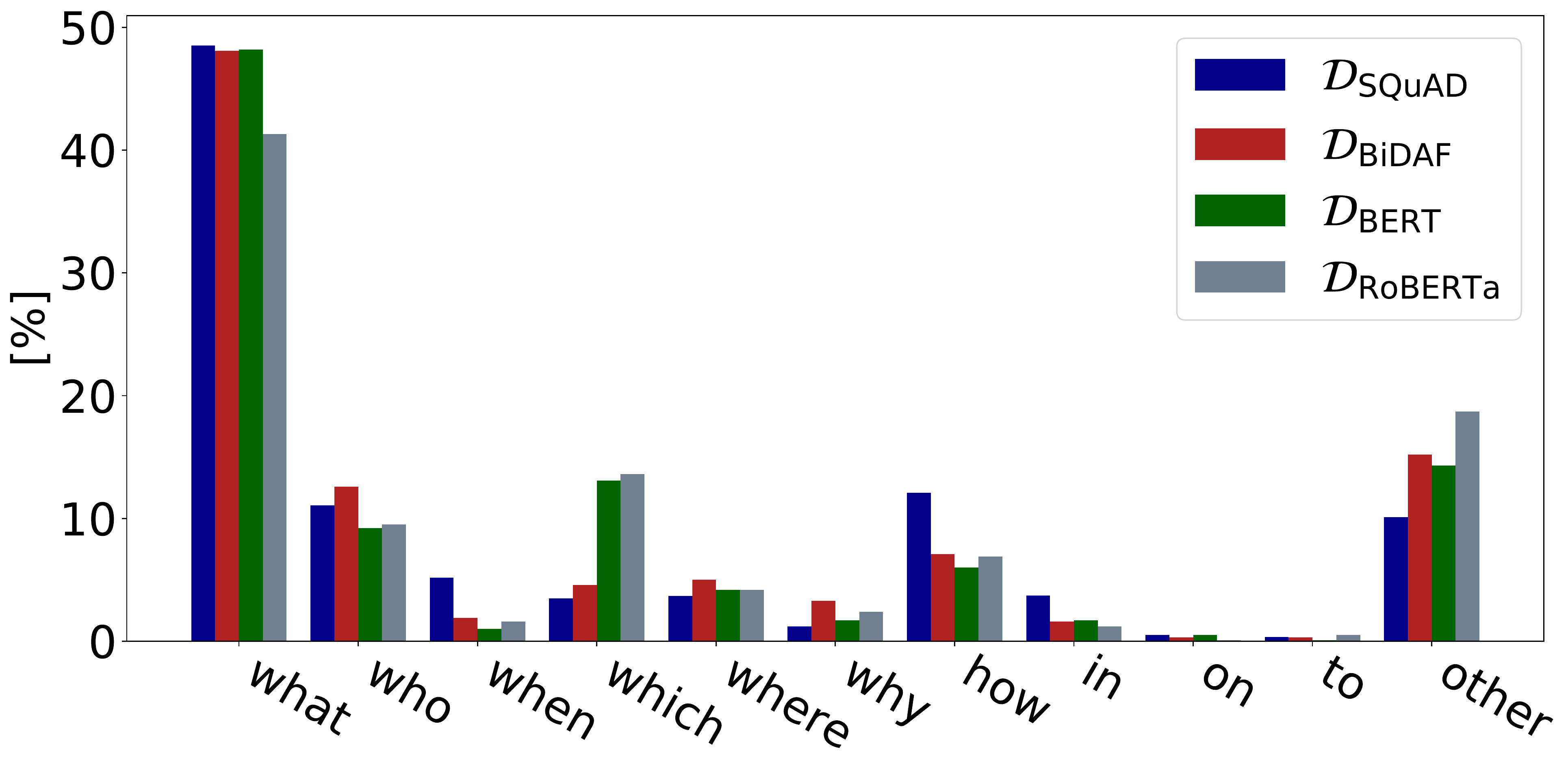}
        \caption{Analysis of question types across datasets.} 
        \label{fig:question_types}
    \end{figure}
    %%%%%%%%%%%%%%%%%%%%%%%%%%%%%%
    
    %%%%%%%%%%%%%%%%%%%%%%%%%%%%%%
    %%%   FIGURE  %%%%
    %%%%%%%%%%%%%%%%%%%%%%%%%%%%%%
    \begin{figure}[t]
        \centering
        \includegraphics[width=\columnwidth]{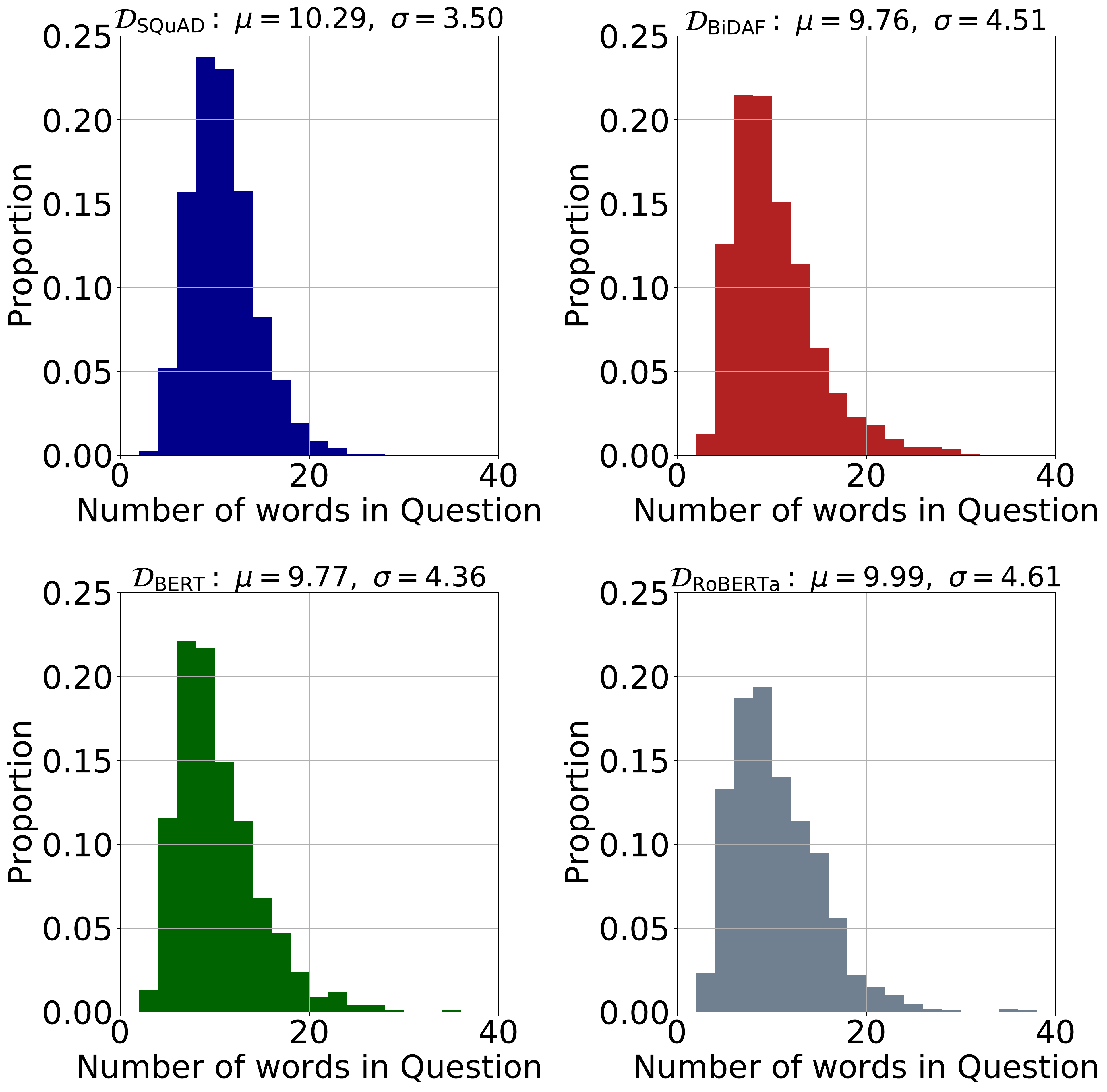}
        \caption{Question length distribution across datasets.} 
        \label{fig:hists_question_len}
    \end{figure}
    %%%%%%%%%%%%%%%%%%%%%%%%%%%%%%
    
    %%%%%%%%%%%%%%%%%%%%%%%%%%%%%%
    %%%   FIGURE  %%%%
    %%%%%%%%%%%%%%%%%%%%%%%%%%%%%%
    \begin{figure}[t]
        \centering
        \includegraphics[width=\columnwidth]{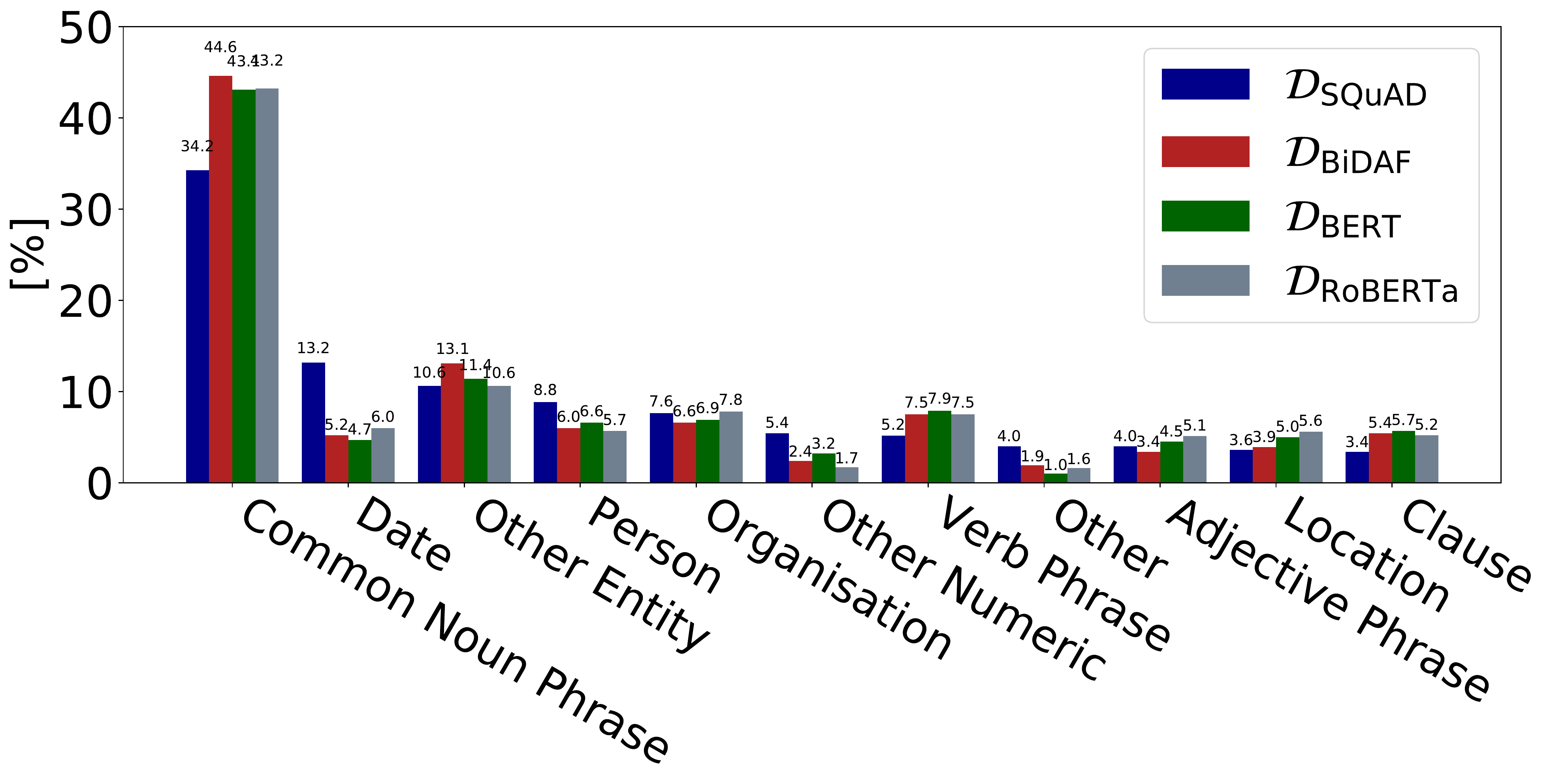}
        \caption{Analysis of answer types across datasets.} 
        \label{fig:answer_types}
    \end{figure}
    %%%%%%%%%%%%%%%%%%%%%%%%%%%%%%
    
    %%%%%%%%%%%%%%%%%%%%%%%%%%%%%%
    %%%   FIGURE  %%%%
    %%%%%%%%%%%%%%%%%%%%%%%%%%%%%%
    \begin{figure}[t]
        \centering
        \includegraphics[width=\columnwidth]{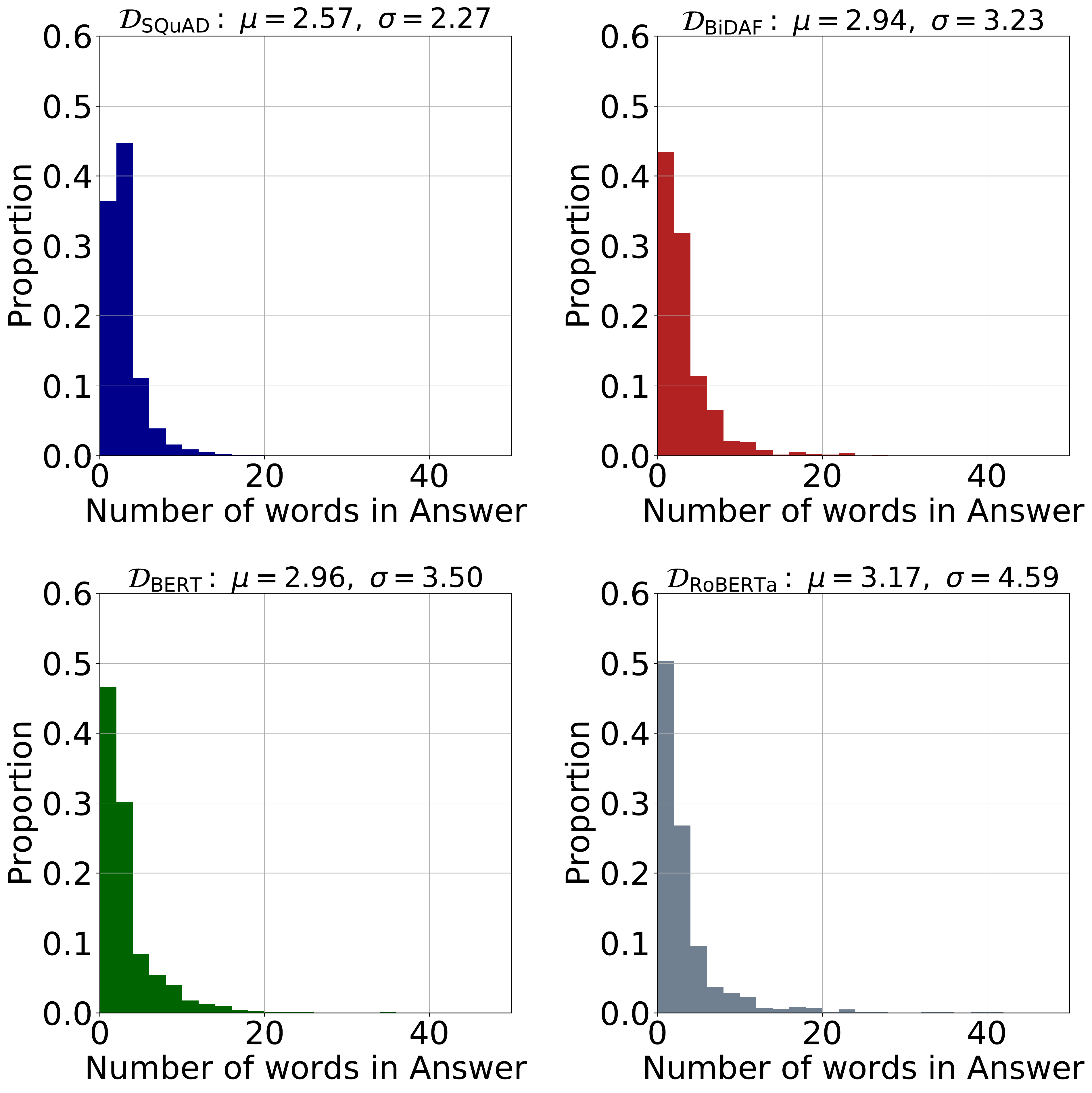}
        \caption{Answer length distribution across datasets.} 
        \label{fig:hists_answer_len}
    \end{figure}
    %%%%%%%%%%%%%%%%%%%%%%%%%%%%%%
    
    \paragraph{Answer statistics}{
    Figure~\ref{fig:hists_answer_len} allows for further analysis of answer lengths across datasets.
    We observe that answers for all datasets constructed with a model in the loop tend to be longer than in \squad{}.
    There is furthermore a trend of increasing answer length and variability with increasing model-in-the-loop strength. 
    We show an analysis of answer types in Figure~\ref{fig:answer_types}.
    }

    %%%%%%%%%%%%%%%%%%%%%%%%%%%%%%
    %%%   Worker Distribution FIGURE  %%%%
    %%%%%%%%%%%%%%%%%%%%%%%%%%%%%%
    \begin{figure*}[t]
        \centering
        \includegraphics[width=\textwidth]{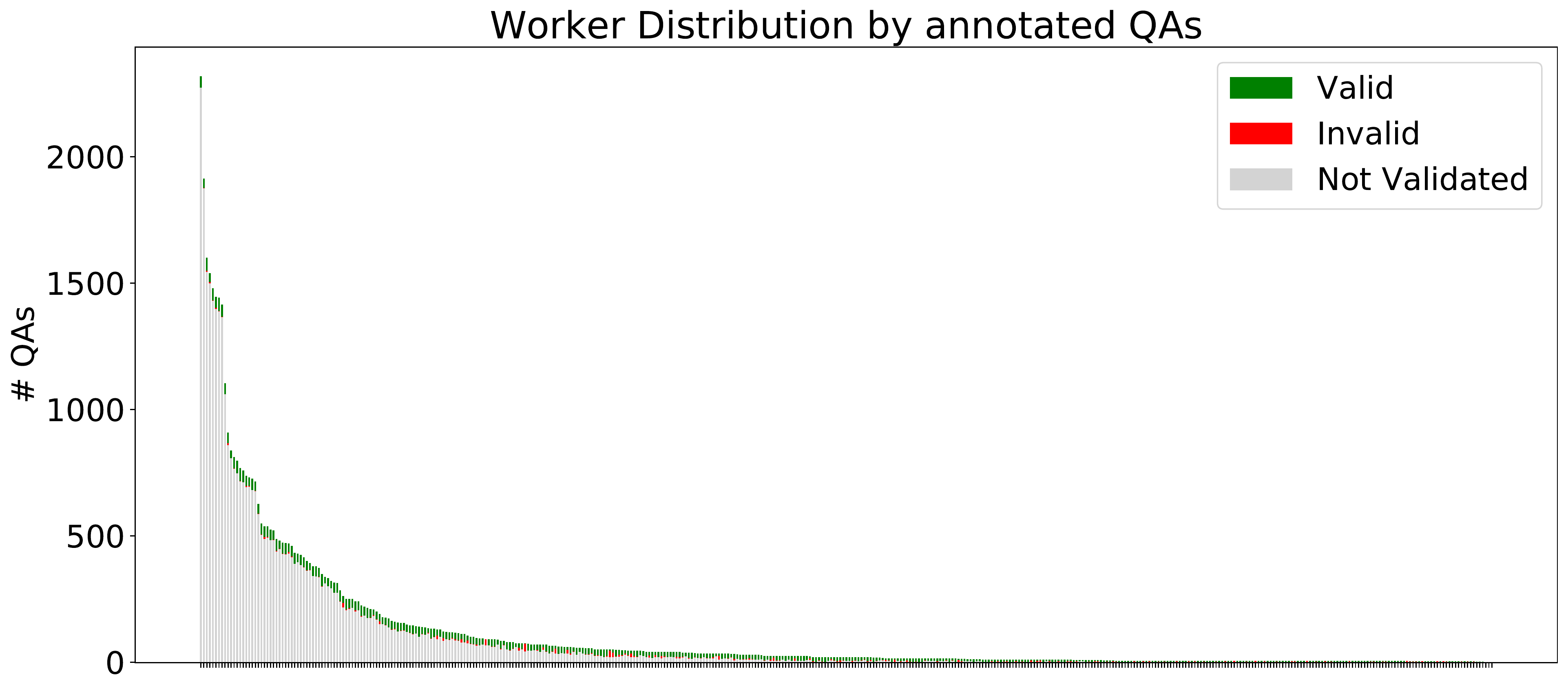}
        \caption{Worker distribution, together with the number of manually validated QA pairs per worker.} 
        \label{fig:worker_distribution}
    \end{figure*}
    %%%%%%%%%%%%%%%%%%%%%%%%%%%%%%
    
    \section{Annotation Interface Details}
    \label{sec:appendix_dataset_construction}
    
    We have three key steps in the dataset construction process: i) training and qualification, ii) ``Beat the AI'' annotation and iii) answer validation.
    
    \paragraph{Training and Qualification}{
    This is a combined training and qualification task; a screenshot of the interface is shown in Figure~\ref{fig:interface_training}.
    The first step involves a set of five assignments requiring the worker to demonstrate an ability to generate questions and indicate answers by highlighting the corresponding spans in the passage. 
    Once complete, the worker is shown a sample ``Beat the AI'' HIT for a pre-determined passage which helps facilitate manual validation. 
    In earlier experiments, these two steps were presented as separate interfaces, however, this created a bottleneck between the two layers of qualification and slowed down annotation considerably. 
    In total, 1,386 workers completed this task with 752 being assigned the qualification.
    }

    \paragraph{``Beat the AI Annotation''}{
    The ``Beat the AI'' question generation HIT presents workers with a randomly selected passage from \squadone{}, about which workers are expected to generate questions and provide answers. 
    This data is sent to the corresponding model-in-the-loop API running on AWS infrastructure and primarily consisting of a load balancer and a \textit{t2.xlarge} EC2 instance with the \textit{T2/T3 Unlimited} setting enabled to allow high sustained CPU performance during annotation runs. 
    The model API returns a prediction which is scored against the worker's answer to determine whether the worker has successfully managed to ``beat'' the model.
    Only questions which the model fails to answer are considered valid; a screenshot for this interface is shown in Figure~\ref{fig:interface_question_generation_full}.
    Workers are tasked to ideally submit at least three valid questions, however fewer are also accepted -- in particular for very short passages. 
    A sample of each worker's HITs is manually validated; 
    those who do not satisfy the question quality requirements have their qualification revoked and all their annotated data discarded. 
    This was the case for 99 workers. 
    Worker validation distributions are shown in Figure~\ref{fig:worker_distribution}.
    }

    \paragraph{Answer Validation}{
    The answer validation interface (cf.~Figure~\ref{fig:interface_validation}) is used to validate the answerability of the validation and test sets for each different model used in the annotation loop. 
    Every previously collected question generation HIT from these dataset parts, which had not been discarded during manual validation, is submitted to at least 3 distinct annotators. 
    Workers are shown the passage and previously generated questions and are asked to highlight the answer in the passage. 
    In a post-processing step, only questions with at least 1 valid matching answer out of 3 are finally retained.
    }
    
    \section{Catalogue of Comprehension Requirements}
    \label{sec:appendix_reasoning_types}
    We give a description for each of the items in our catalogue of comprehension requirements in Table~\ref{tab:reasoning_types}, accompanied with an example for illustration.
    These are the labels used for the qualitative analysis  performed in Section~\ref{sec:qualitative}.

    %%%%%%%%%%%%%%%%%%%%%%%%%%%%%%
    %%%   FIGURE  %%%%
    %%%%%%%%%%%%%%%%%%%%%%%%%%%%%%
    \begin{figure}[ht]
        \centering
        \includegraphics[width=\columnwidth]{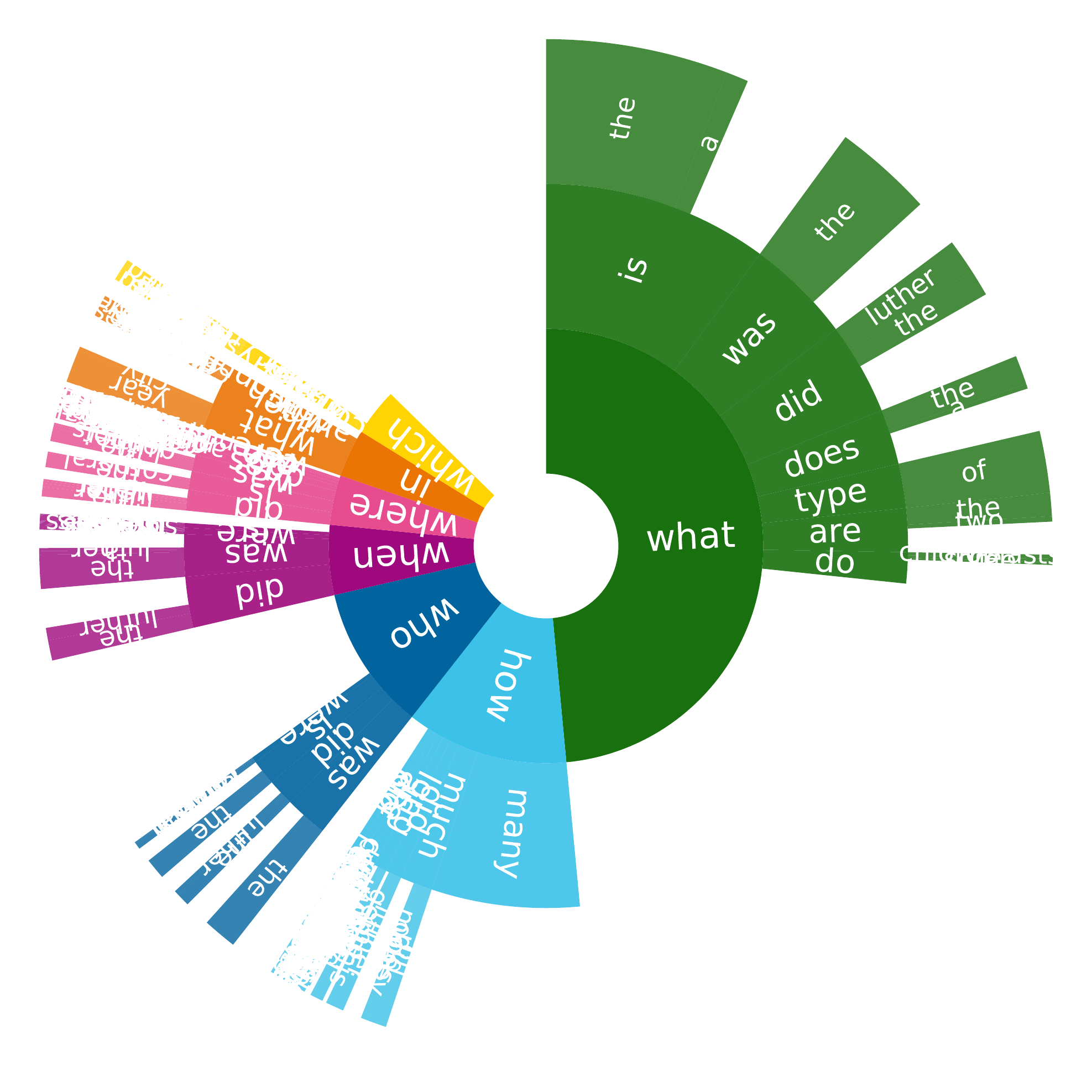}
        \caption{Question sunburst plot for \dataset{SQuAD}.}
        \label{fig:sunburst_squad}
    \end{figure}
    %%%%%%%%%%%%%%%%%%%%%%%%%%%%%%

    %%%%%%%%%%%%%%%%%%%%%%%%%%%%%%
    %%%   FIGURE  %%%%
    %%%%%%%%%%%%%%%%%%%%%%%%%%%%%%
    \begin{figure}[ht]
        \centering
        \includegraphics[width=\columnwidth]{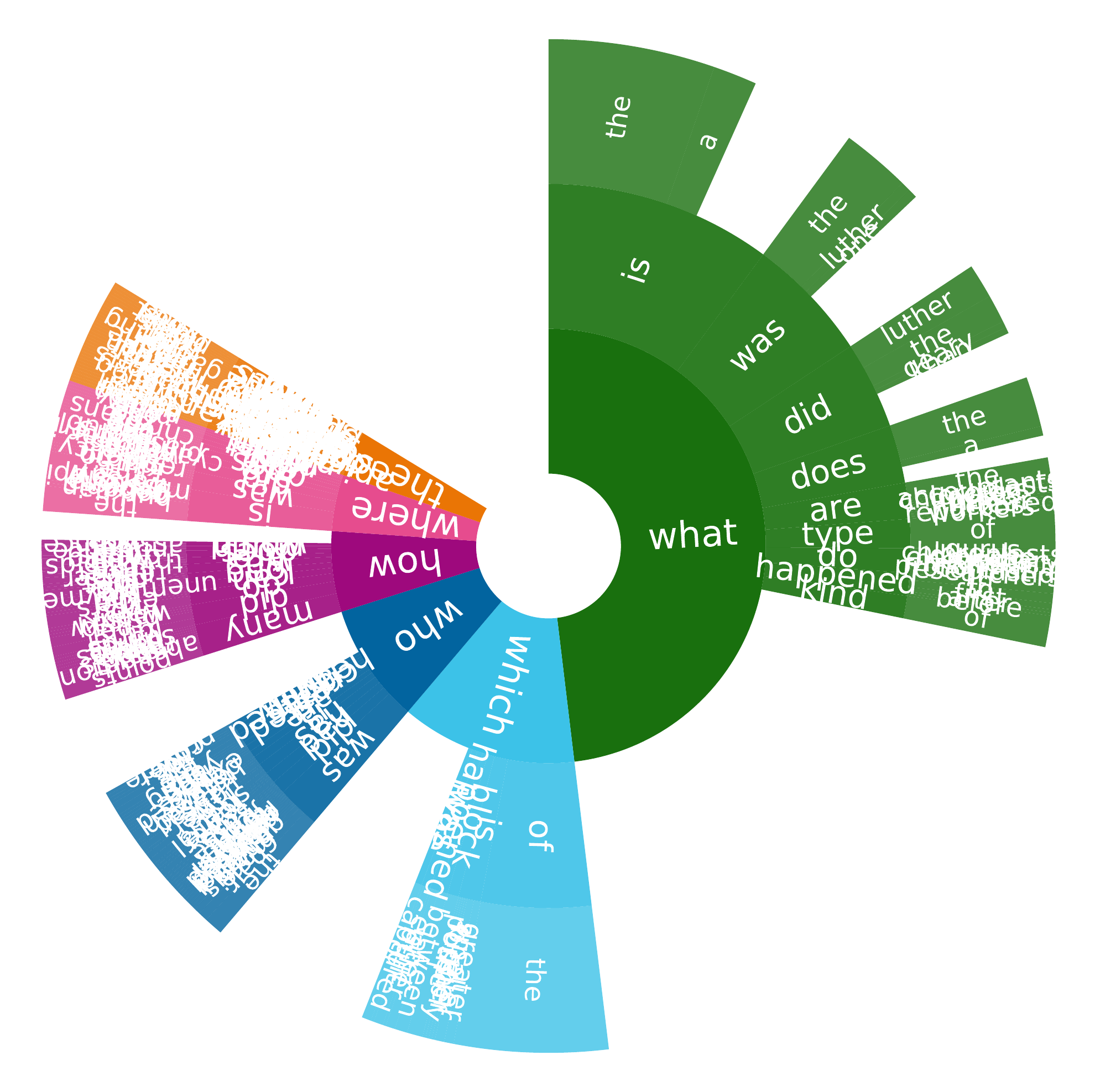}
        \caption{Question sunburst plot for \dataset{BERT}.}
        \label{fig:sunburst_bert}
    \end{figure}
    %%%%%%%%%%%%%%%%%%%%%%%%%%%%%%
    
    %%%%%%%%%%%%%%%%%%%%%%%%%%%%%%
    %%%   FIGURE  %%%%
    %%%%%%%%%%%%%%%%%%%%%%%%%%%%%%
    \begin{figure}[ht]
        \centering
        \includegraphics[width=\columnwidth]{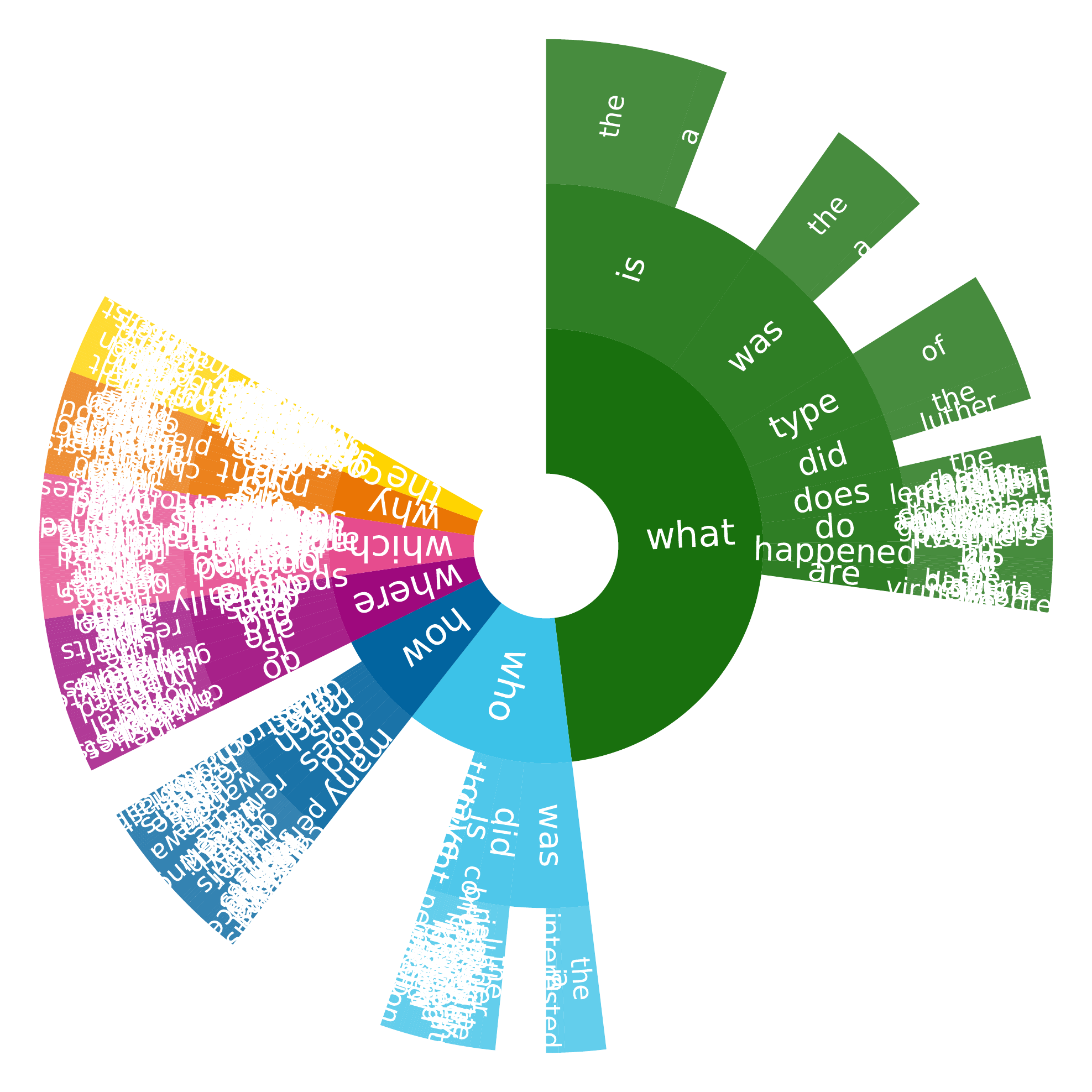}
        \caption{Question sunburst plot for \dataset{BiDAF}.}
        \label{fig:sunburst_bidaf}
    \end{figure}
    %%%%%%%%%%%%%%%%%%%%%%%%%%%%%
    
    %%%%%%%%%%%%%%%%%%%%%%%%%%%%%%
    %%%   FIGURE  %%%%
    %%%%%%%%%%%%%%%%%%%%%%%%%%%%%%
    \begin{figure}[ht]
        \centering
        \includegraphics[width=\columnwidth]{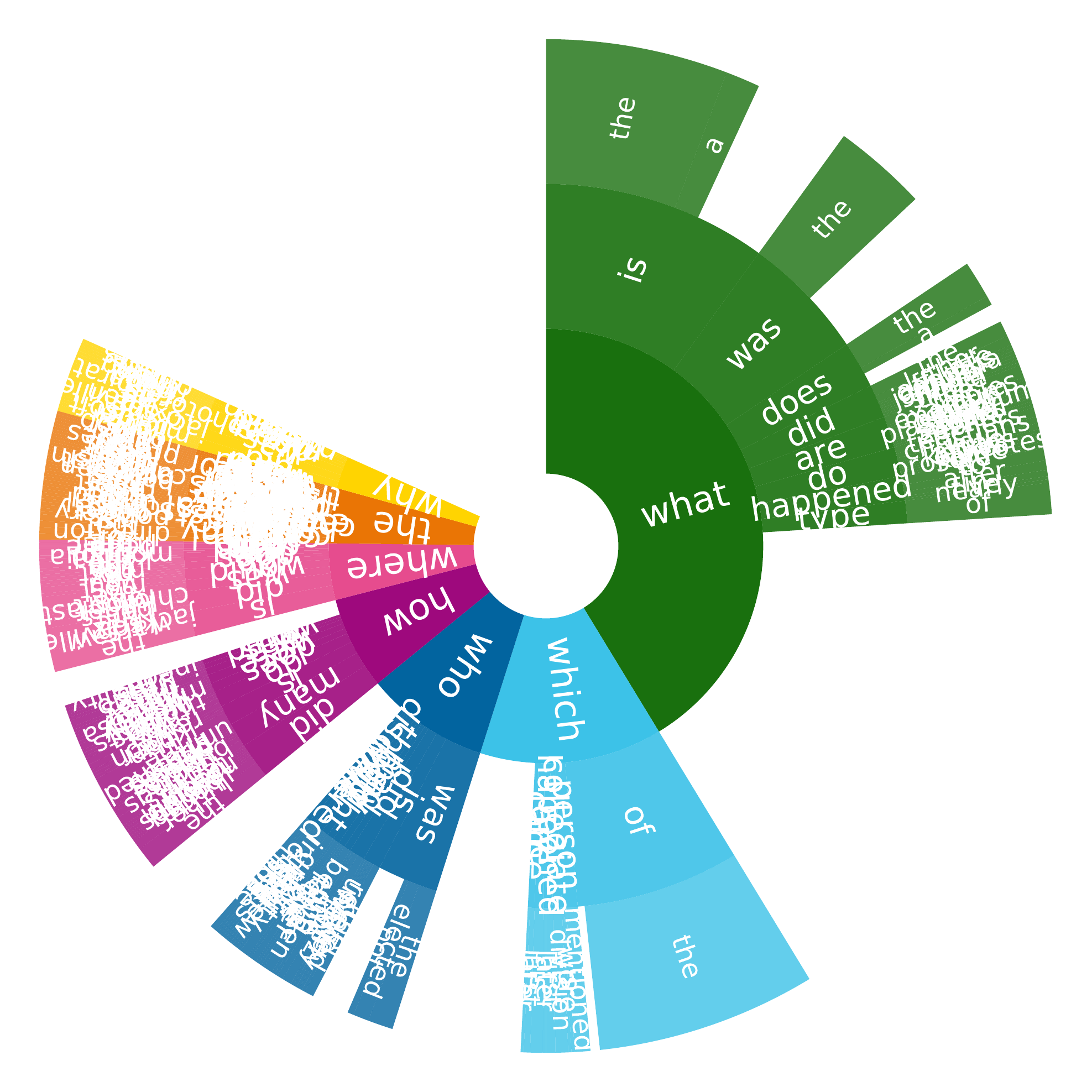}
        \caption{Question sunburst plot for \dataset{RoBERTa}.}
        \label{fig:sunburst_roberta}
    \end{figure}
    %%%%%%%%%%%%%%%%%%%%%%%%%%%%%%
    
    \clearpage

    %%%%%%%%%%%%%%%%%%%%%%%%%%%%%%
    %%%   Training interface FIGURE  %%%%
    %%%%%%%%%%%%%%%%%%%%%%%%%%%%%%
    \begin{figure*}[h]
        \centering
        \includegraphics[width=\textwidth]{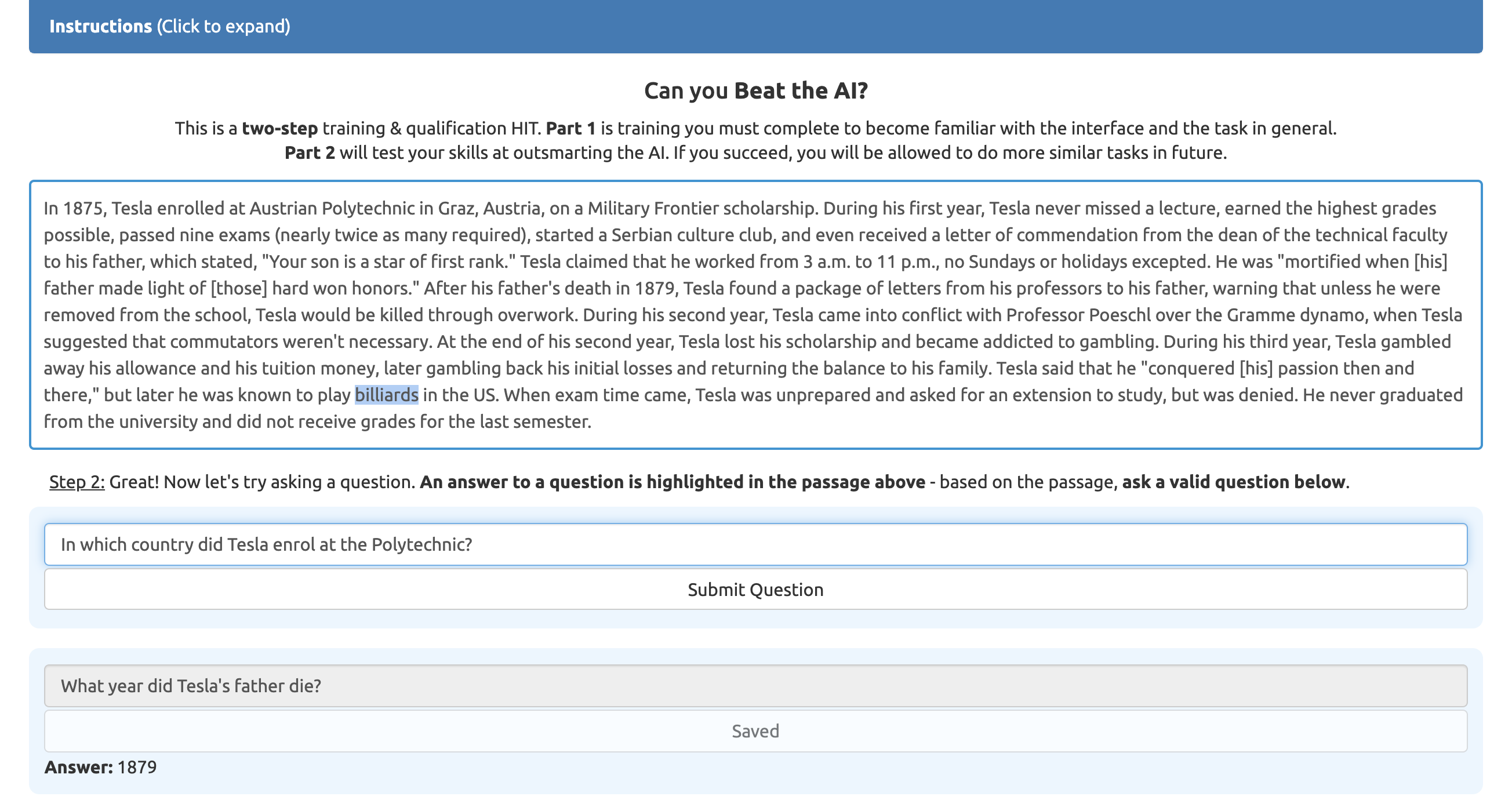}
        \caption{Training and qualification interface. Workers are first expected to familiarise themselves with the interface and them complete a sample ``Beat the AI'' task for validation.} 
        \label{fig:interface_training}
    \end{figure*}
    %%%%%%%%%%%%%%%%%%%%%%%%%%%%%%
    
    %%%%%%%%%%%%%%%%%%%%%%%%%%%%%%
    %%%   Question Generation FIGURE  %%%%
    %%%%%%%%%%%%%%%%%%%%%%%%%%%%%%
    \begin{figure*}[h]
        \centering
        \includegraphics[width=\textwidth]{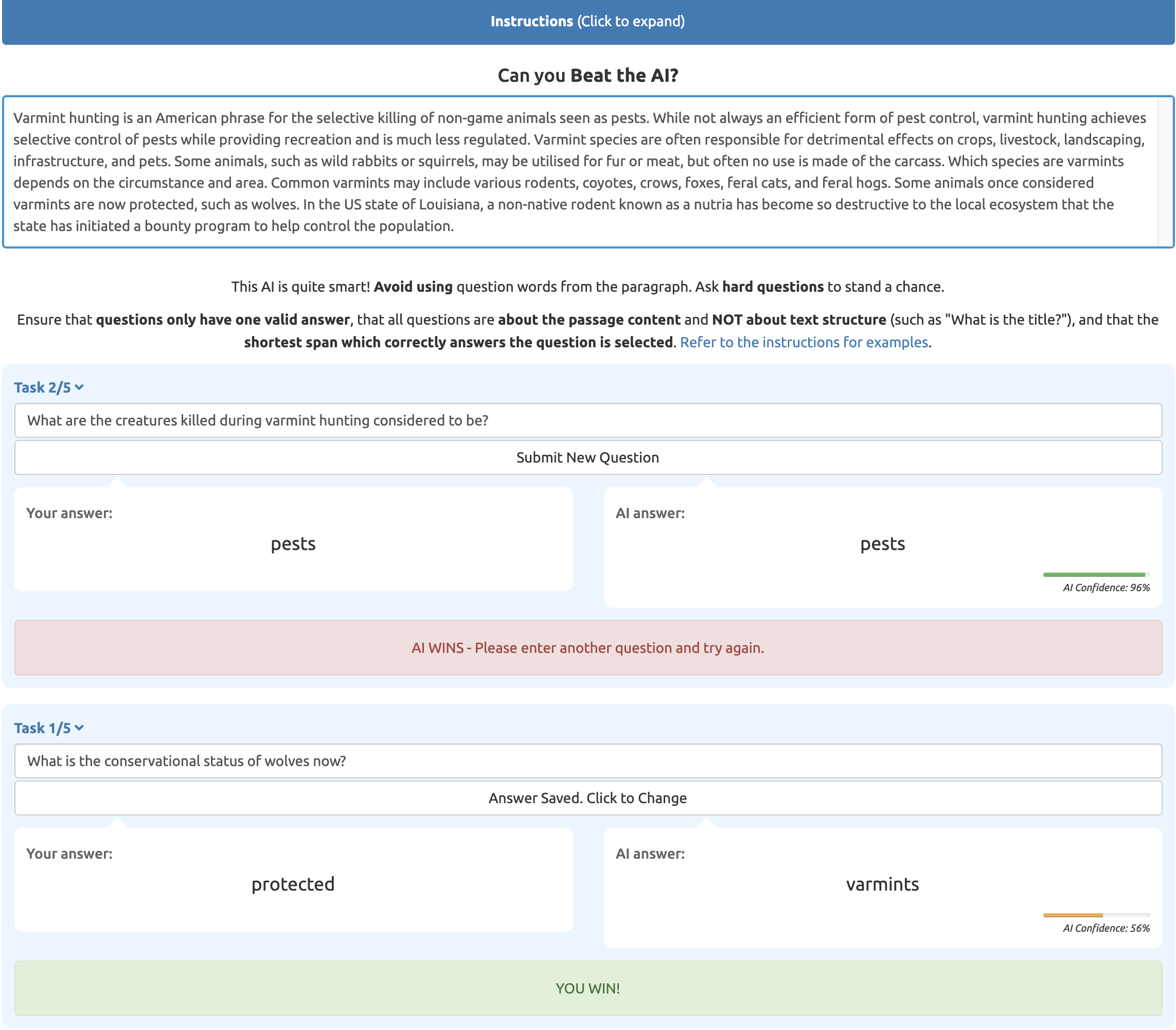}
        \caption{``Beat the AI'' question generation interface. Human annotators are tasked with asking questions about a provided passage which the model in the loop fails to answer correctly.} 
        \label{fig:interface_question_generation_full}
    \end{figure*}
    %%%%%%%%%%%%%%%%%%%%%%%%%%%%%%
    
    %%%%%%%%%%%%%%%%%%%%%%%%%%%%%%
    %%%   Answer Validation interface FIGURE  %%%%
    %%%%%%%%%%%%%%%%%%%%%%%%%%%%%%
    \begin{figure*}[h]
        \centering
        \includegraphics[width=\textwidth]{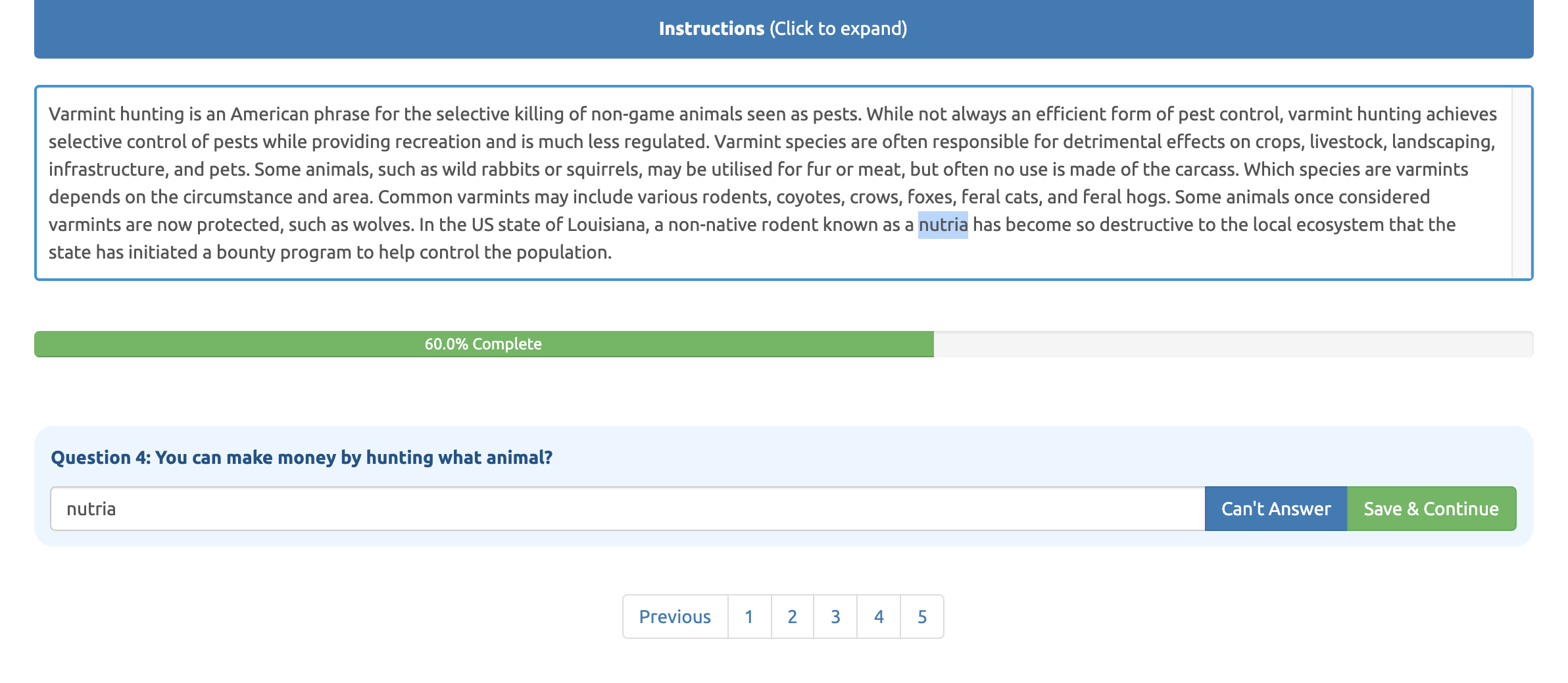}
        \caption{Answer validation interface. Workers are expected to provide answers to questions generated in the ``Beat the AI'' task. The additional answers are used to determine question answerability and non-expert human performance.} 
        \label{fig:interface_validation}
    \end{figure*}
    %%%%%%%%%%%%%%%%%%%%%%%%%%%%%%
    
    \clearpage

    \begin{table*}[ht]
    \footnotesize
    \begin{center}
    \begin{tabular}{p{.1\textwidth}p{.3\textwidth}p{.32\textwidth}p{.18\textwidth}}
    \textbf{Type} & \textbf{Description} &  \textbf{Passage} & \textbf{Question} \\
    \toprule
    Explicit    &  Answer stated nearly word-for-word in the passage as it is in the question.  &  \hlcustom{Sayyid Abul Ala Maududi} was an important early twentieth-century figure in the Islamic revival in India [\ldots]   & Who was an important early figure in the Islamic revival in India? \\
    \hline
    Paraphrasing & Question paraphrases parts of the passage, generally relying on context-specific synonyms. & \hlcustom{Seamans}' establishment of an ad-hoc committee [\ldots]  & Who created the ad-hoc committee? \\
    \hline
    External Knowledge & The question cannot be answered without access to sources of knowledge beyond the passage. & [\ldots] the 1988 film noir thriller Stormy Monday, directed by Mike Figgis and starring Tommy Lee Jones, Melanie Griffith, \hlcustom{Sting} and Sean Bean. & Which musician was featured in the film Stormy Monday? \\
    \hline
    Co-reference & Requires resolution of a relationship between two distinct words referring to the same entity. & Tamara de Lempicka was a famous artist born in Warsaw. [\ldots] Better than anyone else she represented the Art Deco style in \hlcustom{painting and art} [\ldots] & Through what creations did Lempicka express a kind of art popular after WWI? \\
    \hline
    Multi-Hop & Requires more than one step of inference, often across multiple sentences. & [\ldots] and in 1916 married a Polish lawyer Tadeusz \hlcustom{Lempicki}. Better than anyone else she represented the Art Deco style in painting and art [\ldots] & Into what family did the artist who represented the Art Deco style marry? \\
    \hline
    Comparative & Requires a comparison between two or more attributes (e.g.,~\emph{smaller than}, \emph{last}) & The previous chairs were Rajendra K. Pachauri, elected in May 2002; Robert Watson in 1997; and \hlcustom{Bert Bolin} in 1988.  & Who was elected earlier, Robert Watson or Bert Bolin?\\
    \hline
    Numeric & Any numeric reasoning (e.g.,~some form of calculation is required to arrive at the correct answer). &  [\ldots] it has been estimated that \hlcustom{Africans} will make up at least 30\% of the delegates at the 2012 General Conference, and it is also possible that 40\% of the delegates will be from outside [\ldots]  & From which continent is it estimated that members will make up nearly a third of participants in 2012?\\
    \hline
    Negation & Requires interpreting a single or multiple negations. & Subordinate to the General Conference are the \hlcustom{jurisdictional and central conferences} which also meet every four years. & What is not in charge?\\
    \hline
    Filtering & Narrowing down a set of answers to select one by some particular distinguishing feature. & [\ldots] was engaged with Johannes Bugenhagen, Justus Jonas, Johannes Apel, Philipp Melanchthon and \hlcustom{Lucas Cranach the Elder} and his wife as witnesses [\ldots] & Whose partner could testify to the couple's agreement to marry? \\
    \hline
    Temporal & Requires an understanding of time and change, and related aspects. Goes beyond directly stated answers to \emph{When} questions or external knowledge. & In 2010 the Amazon rainforest experienced another \hlcustom{severe drought}, in some ways more extreme than the 2005 drought. & What occurred in 2005 and then again five years later? \\
    \hline
    Spatial & Requires an understanding of the concept of space, location, or proximity. Goes beyond finding directly stated answers to \emph{Where} questions. & Warsaw lies in east-central Poland about 300 km (190 mi) from the Carpathian Mountains and about 260 km (160 mi) from the \hlcustom{Baltic Sea}, 523 km (325 mi) east of Berlin, Germany.  & Is Warsaw closer to the Baltic Sea or Berlin, Germany? \\
    \hline
    Inductive & A particular case is addressed in the passage but inferring the answer requires generalisation to a broader category. & [\ldots] frequently evoked by particular events in his life and the unfolding Reformation. This behavior started with his learning of the \hlcustom{execution} of Johann Esch and Heinrich Voes, the first individuals to be martyred by the Roman Catholic Church for Lutheran views [\ldots] & How did the Roman Catholic Church deal with non-believers? \\
    \hline
    Implicit & Builds on information implied in the passage and does not otherwise require any of the above types of reasoning. & Despite the \hlcustom{disagreements on the Eucharist}, the Marburg Colloquy paved the way for the signing in 1530 of the Augsburg Confession, and for the [\ldots]  & What could not keep the Augsburg confession from being signed? \\
    \bottomrule
    
    \end{tabular}
    \end{center}
    \caption{\label{tab:reasoning_types} Comprehension requirement definitions and examples from adversarial model-in-the-loop annotated RC datasets. Note that these types are not mutually exclusive. The annotated answer is highlighted in yellow.}
    \end{table*}
\fi

\end{document}